\documentclass{article}

 \usepackage[main, final]{neurips_2026}

\usepackage[utf8]{inputenc} 
\usepackage[T1]{fontenc}    
\usepackage{hyperref}       
\usepackage{url}            
\usepackage{booktabs}       
\usepackage{amsfonts}       
\usepackage{nicefrac}       
\usepackage{microtype}      
\usepackage{xcolor}         

\usepackage{graphicx}
\usepackage{algorithm}
\usepackage{algorithmic}
\usepackage{wrapfig}
\usepackage{multirow}

\title{Hit Expansion via Localized Exploration of Synthesizable Chemical Space}

\author{%
  Walter Virany$^{*,1,2,3}$ \quad
  Yidong Jin$^{*,3,4}$ \quad
  Andrew Lian$^{3,5}$ \quad
  Dmytro Shevchuk$^{1,3}$ \\
  \bfseries Amir Mehdi Soufi Enayati$^{3}$ \quad
  Mike Tyers$^{1,3}$ \quad
  Robert A. Batey$^{1,6}$ \quad
  Piotr Gai\'{n}ski$^{7}$ \\
  \bfseries Micha\l{} Koziarski$^{1,2,3}$ \\[.8em]
  $^{1}$University of Toronto \quad
  $^{2}$Vector Institute \quad
  $^{3}$The Hospital for Sick Children \\
  $^{4}$University of Waterloo \quad
  $^{5}$McMaster University \quad
  $^{6}$Acceleration Consortium \\
  $^{7}$Jagiellonian University \quad
  $^*$Equal contribution \\[.5em]
  \texttt{wvirany@cs.toronto.edu, michal.koziarski@sickkids.ca}
}

\newcommand{\std}[1]{{\scriptsize{$\pm$ {#1}}}}

\begin{document}

\maketitle

\begin{abstract}
Generative models for drug discovery which directly produce synthesis routes have gained significant popularity due to their ability to constrain the search space to synthetically accessible molecules. However, existing methods have focused primarily on \textit{de novo} molecular design, and rarely start the generation process from known binders. In this paper, we present HELiX: a template-based GFlowNet for localized exploration of chemical space. HELiX learns to partially decompose a given synthetic trajectory to an intermediate state, and then perform forward synthesis in a manner that preserves synthetic accessibility, leading to diverse, high-reward analog generation. Moreover, we prioritize sample efficiency by incorporating Bayesian optimization into the training procedure. We diagnose problems inherent to training GFlowNets with Bayesian optimization and introduce a greedy acquisition strategy which effectively balances between exploration and exploitation without the need for reward shaping. Finally, we show that local exploration is inherently robust to noisy oracle evaluations, a common problem in drug development when using unreliable proxies for binding affinity.
\end{abstract}

\section{Introduction}
\label{sec:introduction}

In recent years, generative models for molecular design have attracted substantial attention in the machine learning community \citep{tong2021generative,van2025search}, providing a powerful alternative to traditional supervised approaches for exploring vast chemical spaces. By learning to sample molecules from a target distribution, these models can directly propose candidates with desired properties, effectively acting as data-driven molecular search engines. While early models imposed few constraints on the structures they produced, the field has recently shifted toward incorporating molecular synthesizability to enable the direct translation of \textit{in silico} designs to wet-lab experiments \citep{koziarski2024,cretu2025synflownet,gainski2025scalable}, a consideration that has become increasingly important with the widespread adoption of iterative design–make–test–analyze (DMTA) pipelines in drug discovery \citep{tom2024self}.

Despite their growing impact, most existing generative models—both unconstrained and synthesizability-aware—primarily focus on \textit{de novo} molecular design, generating compounds from scratch without reference to an existing molecule. While \textit{de novo} design is an important component of molecular discovery, later stages such as hit expansion and molecular optimization are equally critical, pose distinct challenges, and have received comparatively little attention. In particular, these tasks require localized exploration around a given starting point and place strong demands on sample efficiency: oracle evaluations in later-stage campaigns are expensive, budgets are limited, and the cost of each experiment grows as candidates advance through the pipeline.

\begin{figure*}[t]
    \centering
    \includegraphics[width=0.9\linewidth]{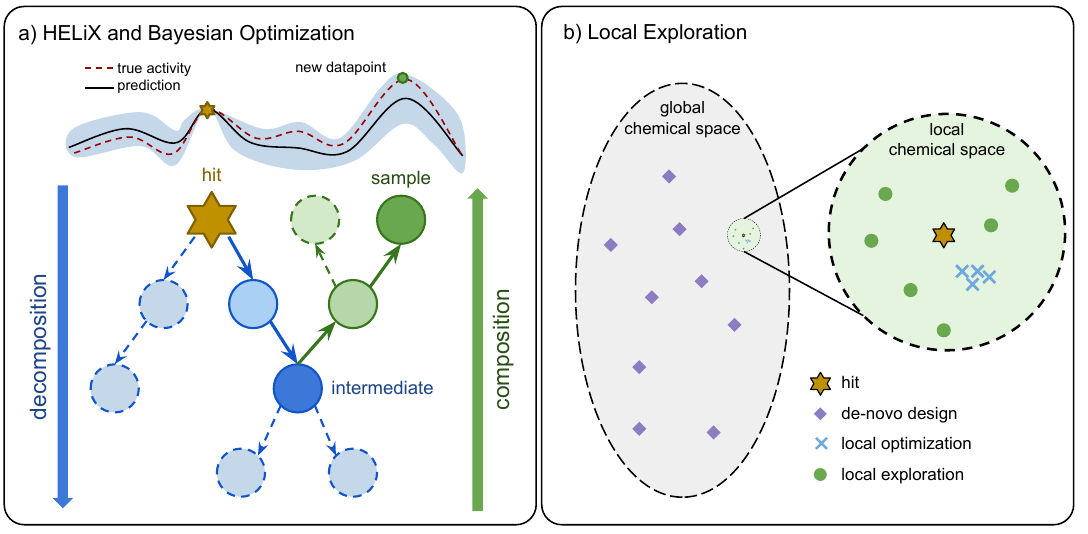}
    \caption{(a) \textbf{The HELiX generation process.} A given hit molecule is decomposed by reversing reaction steps to an intermediate, which is then recomposed into a new sample. Generated candidates are evaluated by an oracle within a Bayesian optimization loop, and observations are used to update the surrogate model guiding subsequent sampling. (b) \textbf{Local exploration.} Unlike \textit{de novo} design, which performs global exploration of chemical space, HELiX concentrates sampling in the local chemical neighborhood of a known hit while preserving shared synthesis routes.}
    \label{fig:helix_overview}
    \vspace{-6pt}
\end{figure*}

To address this gap, we introduce HELiX, a framework for \textbf{h}it \textbf{e}xpansion via \textbf{l}ocalized e\textbf{x}ploration of synthesizable chemical space. HELiX is, to our knowledge, the first generative model that enables localized exploration around a provided hit while inheriting both guaranteed synthesizability and diversity from proportional reward sampling. Our contributions are listed as follows:

\begin{itemize}
    \item We propose HELiX, an extension of template-based generative flow networks (GFlowNets) that reformulates the underlying Markov decision process (MDP) to support reversible reaction steps, enabling controlled local exploration (Figure~\ref{fig:helix_overview}).
    \item We improve sample efficiency of the framework by integrating Bayesian optimization (BO) into GFlowNet training, diagnosing and addressing failure modes that arise when training GFlowNets with standard BO approaches. Furthermore, we introduce a greedy acquisition strategy that balances exploration and exploitation in GFlowNet sampling without the need for reward shaping.
    \item We present an extensive experimental evaluation, demonstrating that HELiX achieves state-of-the-art performance in localized chemical exploration, exhibiting high diversity, strong sample efficiency, and robustness to noisy oracle evaluations, while guaranteeing synthesizability out of the box.
\end{itemize}

\section{Background and Related Work}
\label{sec:gen_inst}

\subsection{GFlowNets}
\label{sec:background_gflownets}

GFlowNets \citep{bengio2021,bengio2023} are probabilistic models that learn a stochastic policy for constructing objects $x$ through sequences of actions, such that the probability of sampling an object is proportional to a predefined reward $R(x)$. Generation is formulated as a trajectory $\tau \in \mathcal{T}$ starting from an initial state $s_0$, transitioning between states via actions $a$.

GFlowNets define a forward policy $P_F(s' \mid s)$ and a backward policy $P_B(s \mid s')$, trained using the Trajectory Balance objective \citep{malkin2022}:
\begin{equation}
    Z\prod_{t=0}^{T-1} P_F(s_{t+1} \mid s_t) = R(x)\prod_{t=0}^{T-1} P_B(s_t \mid s_{t+1}),
\end{equation}
where $Z$ is the total flow through the source state and $x$ is the terminal state of the trajectory. GFlowNets have been applied to a broad range of scientific discovery problems, including molecular design \citep{koziarski2024,kim2024,cretu2025synflownet,shen2024}, biological sequence design \citep{jain_biological_2023,ghari2024}, retrosynthesis \citep{gainski2025diverse}, molecular conformation sampling \citep{volokhova2024towards}, and crystal generation 
\citep{ai4science2023crystal}.

\subsection{Synthesizable Molecular Generation}
\label{sec:background_synth}

Ensuring the synthesizability of generated molecules has been approached in several ways. One common strategy relies on synthesizability scoring functions, either as optimization objectives or post-hoc filters \citep{korablyov2024generative}. These include heuristic metrics \citep{ertl2009estimation, genheden2020aizynthfinder} and machine-learning-based predictors \citep{liu2022retrognn, guo2025directly}. However, such scores can be unreliable, particularly out of distribution, and often fail to capture practical considerations such as synthesis cost or reaction availability.

As an alternative, many methods explicitly constrain generation to chemically feasible spaces by constructing molecules from predefined building blocks and reaction templates, referred to as template-based approaches \citep{bradshaw2019model,bradshaw2020barking,gao2021amortized,gottipati2020learning,horwood2020molecular,seo2023molecular,gao2024synformer,jocys2024synthformer,rekesh2025syncogen,lo2025synga,lee_exploring_2026}. Template-based GFlowNets have recently emerged as a powerful paradigm for molecular generation in drug design, combining the diversity guarantees of flow-based objectives with the ability to generate molecules with fully specified synthesis routes \citep{koziarski2024,cretu2025synflownet,seo2025generative,gainski2025scalable}.

\subsection{Hit Expansion}
\label{sec:background_hit_expansion}

Most computational drug design methods frame molecule generation as a global optimization problem, aiming to directly maximize predicted properties such as binding affinity \citep{GomezBombarelli2018,Olivecrona2017,You2018,Stahl2019}. In contrast, hit expansion requires the generation of structurally proximal analogs around known binders, aligning more closely with medicinal chemistry practice and structure-activity relationship development. Existing generative approaches differ in how proximity to the initial hit is enforced. SynFormer-ED \citep{gao2024synformer} conditions a decoder on an encoding of the hit, so proximity is encouraged by conditioning but is neither guaranteed nor directly controllable. REINVENT \citep{loeffler2024reinvent} achieves locality from a prior trained on similar molecule pairs, but places no constraints on synthesizability. SynGA \citep{lo2025synga} operates directly on synthesis trees, obtaining locality through the fitness function or the initial population rather than through the generative process. Projection methods map a reference molecule onto a nearby synthesizable pathway \citep{luo2024projecting}. ReaSyn \citep{lee_exploring_2026} further guides this with oracle feedback. We highlight that most of these methods were not developed for oracle-guided local optimization, and required adaptations in our baseline comparisons, which we discuss in Appendix~\ref{sec:baselines}.

HELiX enforces locality by design: a generated molecule must share a partial synthesis route with the hit by construction. This process is directly controllable through the number of decomposition and forward reactions steps rather than implicit. Other synthesis-aware GFlowNets \citep{koziarski2024,cretu2025synflownet,gainski2025scalable,kim2026synthesizablemoleculargenerationsoftconstrained,zhu_syngfn_2026} generate molecules unconditionally from an empty initial state and are not designed to expand a given hit. To our knowledge, HELiX is the first synthesis-aware GFlowNet to perform local hit expansion.

\subsection{Sample Efficiency in GFlowNets}
\label{sec:background_sample_eff}
Sample efficiency is a central challenge for GFlowNets in molecular design, as training typically requires many oracle evaluations to converge \citep{pmo}. This is particularly problematic in later-stage drug discovery campaigns, where 
experimental validation is costly and oracle budgets are severely constrained. Prior work aims to address this 
through improved training objectives such as Trajectory Balance \citep{malkin2022} and hybrid evolutionary schemes \citep{kim2024}.

BO offers a complementary approach by adaptively selecting the most informative candidates for oracle evaluation under a limited budget \citep{shahriari2015taking,garnett_bayesoptbook_2023}. BO has been widely adopted in molecular design and active learning settings [\citealp{GomezBombarelli2018}, \mbox{\citealp{tripp2023genetic}}, \citealp{korablyov2024generative}], and has been explored in conjunction with GFlowNets \citep{jain_biological_2023, hernandez-garcia_mfal_2024}. However, because GFlowNets sample proportionally to a reward function rather than maximizing it, combining them with standard BO acquisition functions introduces non-trivial challenges: the shape of the acquisition landscape matters, not only the location of its maximum. In this work, we study these challenges in detail and provide strategies to overcome them.

\section{Methods}
\label{sec:methods}

In this paper, we propose HELiX, a template-based GFlowNet framework for local exploration of chemical space. Similarly to other template-based GFlowNets, HELiX operates on a predefined set of reaction templates and building blocks. Below, we describe the formulation of the MDP underlying HELiX, and how BO is incorporated to achieve sample-efficient training.

\subsection{Forward and Backward Policies}
\label{sec:forward_and_backward_policies}

Given a synthetic route to an initial hit, HELiX decomposes reaction steps until reaching an intermediate molecule. From this intermediate, a composition phase applies forward reactions resulting in a new molecule. The number of decomposition steps and forward reaction budget are specified as hyperparameters (see Section~\ref{sec:experimental_details}). States are represented as a tuple $(m, r, F)$, where $m$ is the current molecule, $r$ is a reaction template, and $F$ is a list of fragments. The reaction template $r$ may be unset (denoted by $\emptyset$), and the fragment list $F$ may be empty or partially filled. Moreover, we track and limit the number of decomposition and composition steps. We describe each specific action space in more detail below.

\paragraph{Decomposition Forward Action Space.}
Decomposition begins from the hit molecule $H$ in the initial state $s_0 = (H, \emptyset, \emptyset)$. All reaction templates $r$ for which $H$ is a valid product are enumerated, and the policy model selects one of them. The corresponding set of reactants then contains fragments $F$ and a non-fragment molecule $m$, yielding the state $(m, r, F)$ such that applying reaction $r$ to fragments $F$ and molecule $m$ yields $H$. The fragments in $F$ are then discarded one by one until reaching a state of the form $(m, r, \emptyset)$. At this point, the reaction template $r$ can be removed, resulting in the state $(m, \emptyset, \emptyset)$, from which further decomposition actions may be taken. The decomposition process may terminate at any time, yielding a partially specified state of the form $(m, *, *)$.

\paragraph{Composition Forward Action Space.}
Starting from an intermediate state $(m, *, *)$, the composition process proceeds by sequentially selecting a reaction template $r$ and any required fragments $F$, and applying the resulting reaction $(m, r, F)$ to obtain a new state $(m', \emptyset, \emptyset)$. From this complete state, the agent may either continue composing by selecting another template and fragments, or terminate the process, producing a final molecule. The number of composition steps is bounded by a budget that increases with each decomposition step taken, ensuring the final molecule remains structurally proximate to the hit (see Section~\ref{sec:experimental_details}).

\paragraph{Backward Action Space.}
Traversing the MDP backward introduces two main challenges. First, when reversing the composition phase, the trajectory must end at an intermediate molecule from which the initial hit can be recovered by reversing the decomposition phase. Second, when reversing decomposition, only molecules that lie on a valid synthesis path of the hit are admissible. Explicitly enforcing the validity of all backward actions along a trajectory is prohibitively expensive---more so than the retrosynthesis problem considered in \citep{koziarski2024}. Instead, following \citep{cretu2025synflownet}, we encourage backward policy $P_B$ of HELiX to assign high probability only to actions that could eventually lead to a hit molecule by optimizing it directly with the REINFORCE algorithm \citep{williams1992simple}:
\begin{equation}
    J_B = E_{\tau \sim P_B} [R_{B}(\tau)] - \alpha \mathcal{H}[P_B]
\end{equation}
where $\mathcal{H}[P_B]$ is the entropy term and the reward $R_B(\tau)$ is set to $1$ if trajectory $\tau$ ends in the hit, and $-1$ otherwise.

\subsection{Bayesian Optimization in GFlowNets}
\label{sec:methods_bo}

We incorporate BO into our GFlowNet training procedure by using standard acquisition functions directly as the reward function (see Section~\ref{sec:appendix_bo} for details on the surrogate model and acquisition function definitions).

\begin{figure}[t]
    \centering
    \includegraphics[width=.8\linewidth]{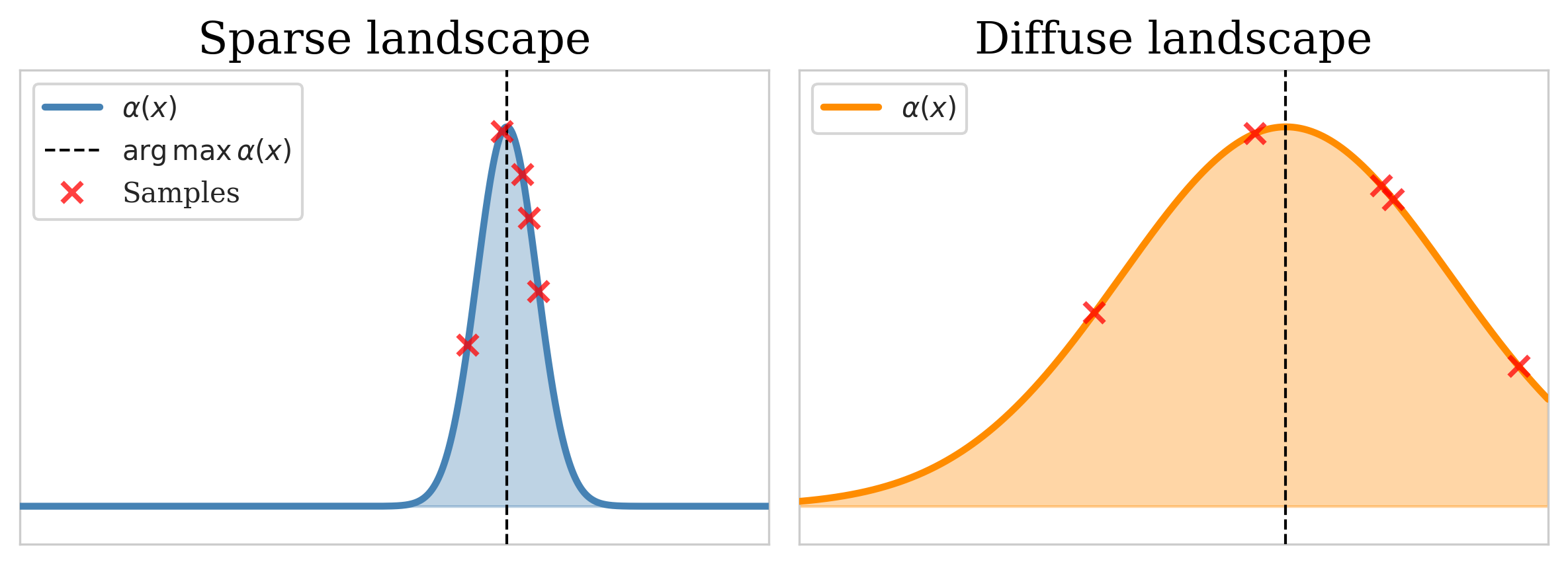}
    \caption{Behavior of proportional sampling versus maximization under sparse (left) and diffuse (right) acquisition landscapes. Under a sparse landscape, proportional sampling concentrates probability mass on high-reward regions, approaching the behavior of maximization in the limit of a Dirac delta distribution.}
    \label{fig:sampling_schematic}
    \vspace{-6pt}
\end{figure}

\paragraph{Sampling acquisition functions}
Unlike standard BO, where candidates are selected by directly maximizing the acquisition function $\alpha(x)$, GFlowNets generate candidates by sampling proportionally to it. This distinction results in dramatically different behavior for various standard acquisition functions such as Expected Improvement (EI) and Upper Confidence Bound (UCB), where the \textit{shape} of the acquisition landscape matters, not only the location of its maximum. We contrast the behavior of maximization and proportional sampling under different acquisition landscapes in Figure~\ref{fig:sampling_schematic}. A sparse landscape---one in which $\alpha(x) \approx 0$ for most candidates in the design space $\mathcal{X}$---will naturally result in exploitative behavior, where the probability mass for sampling is concentrated in high-reward regions. Alternatively, for diffuse acquisition landscapes, sampling will result in wider exploration. Beyond its effect on sampling behavior, landscape sparsity also poses challenges for GFlowNet training: when $\alpha(x) \approx 0$ for most of $\mathcal{X}$, the trajectory balance objective receives little signal to differentiate between candidates, impeding convergence. We study this in more detail in Section~\ref{sec:experiments_bo}.

\paragraph{Reward shaping}
A standard approach in GFlowNet training to recover more exploitative behavior during sampling is reward shaping, in which the reward function is transformed to concentrate on high-reward regions \citep{bengio2021}. Specifically, we apply a power transformation to the observations $\tilde{y} = y^\beta$ prior to fitting the surrogate model, where $\beta > 1$. As $\beta$ increases, the acquisition landscape becomes increasingly concentrated in regions with high observed values. However, this can impact the diversity benefits of proportional sampling, and in practice introduces a sensitive hyperparameter tuning problem.

\paragraph{Greedy subsampling}
As an alternative to modifying the reward landscape, we introduce a greedy acquisition subsampling approach. We use the GFlowNet to generate a pool of $M$ candidates, then apply a secondary subsampling step to select the final batch of $q < M$ candidates for evaluation. This approach circumvents the need for careful hyperparameter tuning of the reward transformation, and allows the use of smooth reward landscapes, resulting in more stable GFlowNet training and diverse candidate generation. Moreover, it allows for the use of batch acquisition functions over the fixed pool of generated candidates, which otherwise cannot be used directly to train the GFlowNet due to the non-stationary reward signal introduced by batch dependency. We consider several choices of subsampling methods and study the effect of the generated pool size $M$ in Section~\ref{sec:experiments_bo}.

\section{Experiments}
\label{sec:experiments}

\subsection{Comparison to Baseline Methods for Local Exploration}
\label{sec:experiments_comparison_to_baselines}

\paragraph{Baselines.}
We compare HELiX against state-of-the-art baselines for local hit expansion, including SynFormer-ED \citep{gao2024synformer}, REINVENT4 \citep{loeffler2024reinvent}, TRACER \citep{nakamura2025tracer}, SynGA \citep{lo2025synga}, and ReaSyn \citep{lee_exploring_2026}. For comparison, we also include SCENT \citep{gainski2025scalable}, a template-based GFlowNet designed for \textit{de novo} drug design, as well as a uniform policy baseline, which samples directly from the HELiX state space without a learned policy.

\paragraph{Evaluation  metrics.}
We evaluate each method's ability to perform local exploration by restricting evaluation to \textit{analogs} of the initial hit. For all experiments, we define an analog to be within 0.4 ECFP4 Tanimoto similarity to the initial hit. Reported metrics include analog rate and success rate, measuring the proportion of total molecules identified as analogs and the proportion of molecules that are identified as analogs \textit{and} improve upon the initial hit. Additionally, we report the number of high-reward modes and Murcko scaffolds discovered, as well as the top-1 and top-10 mean reward. Modes are defined as molecules in the observed set above a given reward threshold with < .7 pairwise Tanimoto similarity to other modes. We discuss the analog definition in Section~\ref{sec:ablations_sim}, where we analyze the chemical significance of the threshold and show that our results are not sensitive to alternative definitions of similarity. We emphasize that this criterion is used only for evaluation: no term in the HELiX reward depends on similarity to the hit, so the analog metrics are held out with respect to the model. Locality in HELiX follows entirely from the shared synthesis route.

\paragraph{Oracles.}
We evaluate all methods using pre-trained proxy models for binding affinity to three therapeutically relevant protein targets: sEH proxy \citep{bengio2021}, as well as two oracles from the Therapeutics Data Commons (TDC): JNK3 and GSK3$\beta$ \citep{tdc}. We report results on sEH in this section and defer the results for JNK3 and GSK3$\beta$ to Section~\ref{sec:additional_results}. In Section~\ref{sec:experiments_virtual_screening}, we additionally use AutoDock Vina  as an oracle for virtual screening \citep{vina2}.

\paragraph{Results.}
Results for models trained with 10k oracle calls are shown in Table~\ref{tab:baselines_seh}. HELiX achieves the highest success rate by a wide margin, identifying analogs that improve upon the initial hit at a rate of 11\%. Notably, REINVENT4 discovers the highest analog rate at 87\% among optimization methods, yet almost none of these improve upon the hit, highlighting that analog discovery alone is insufficient for efficiently navigating the local neighborhood of the hit. HELiX also achieves the highest number of high-reward modes and scaffolds, demonstrating that it identifies a diverse set of high-quality analogs rather than concentrating on a single region. In addition, HELiX outperforms SynGA; we highlight that genetic algorithms are generally strong baselines for molecular optimization, particularly in terms of sample efficiency \citep{tripp2023genetic}. The poor performance of SCENT relative to HELiX highlights the necessity of local exploration over \textit{de novo} methods. Finally, the uniform policy baseline demonstrates that the performance of HELiX is not merely an artifact of the restricted state space, a non-trivially large space as shown in Section~\ref{sec:ablations_library}, but requires a learned policy to navigate effectively. We report additional results at an alternative reward threshold (proxy score $> 8$) in Section~\ref{app:seh_threshold}.

\begin{table*}[t]
\caption{Comparison to baselines on the sEH proxy. All metrics are computed over analogs of the initial hit, defined as molecules within .4 ECFP4 Tanimoto similarity. Analog rate denotes the proportion of total molecules identified as analogs; success rate denotes the proportion of analogs that improve upon the initial hit. Modes and scaffolds are counted among high-reward analogs (proxy score $> 7.2$) with pairwise Tanimoto similarity below 0.7. Values are reported as mean $\pm$ standard deviation across initial hits; the number of hits evaluated per method is given in Appendix~\ref{sec:coverage}.}
\label{tab:baselines_seh}
\vspace{6pt}
\centering
\begin{tabular}{lcccccc}
\toprule
\textbf{Method} & \textbf{Analog Rate} & \textbf{Success Rate} & \textbf{Modes} & \textbf{Scaffolds} & \textbf{Top-1} & \textbf{Top-10} \\
\midrule
Uniform     & \textbf{.91} \std{.04} & .006 \std{.003} & 166 \std{19} & 461 \std{98} & 8.12 \std{.07} & 7.97 \std{.03} \\
SCENT              & .07 \std{.07} & .002 \std{.002} & 174 \std{160} & 613 \std{610} & 7.45 \std{2.5} & 7.40 \std{2.4} \\
SynFormer       & .65 \std{.14} & .06 \std{.05}  & 549 \std{126} & 3417 \std{687} & 8.30 \std{.09} & \textbf{8.24} \std{.08} \\
REINVENT        & .87 \std{.08} & .003 \std{.004} & 203 \std{80} & 418 \std{191} & 8.09 \std{.08} & 7.99 \std{.09} \\
TRACER             & .54 \std{.20} & .002 \std{.003} & 92 \std{64}  & 243 \std{170} & 8.04 \std{.15} & 7.84 \std{.27} \\
SynGA     & .33 \std{.1} & .007 \std{.008} & 194 \std{55} & 810 \std{317} & 8.11 \std{.06} & 8.19 \std{.04} \\
ReaSyn     & .3 \std{.2} & .001 \std{.001} & 66 \std{66} & 379 \std{534} & 7.96 \std{.3} & 7.84 \std{.29} \\
HELiX        & .65 \std{.17} & \textbf{.11} \std{.08} & \textbf{933} \std{227} & \textbf{4559} \std{1221} & \textbf{8.31} \std{.07} & 8.22 \std{.05} \\
\bottomrule
\end{tabular}
\vspace{-8pt}
\end{table*}

\subsection{Sample Efficiency}
\label{sec:sample_efficiency}
\begin{figure}[h]
    \centering
    \includegraphics[width=\linewidth]{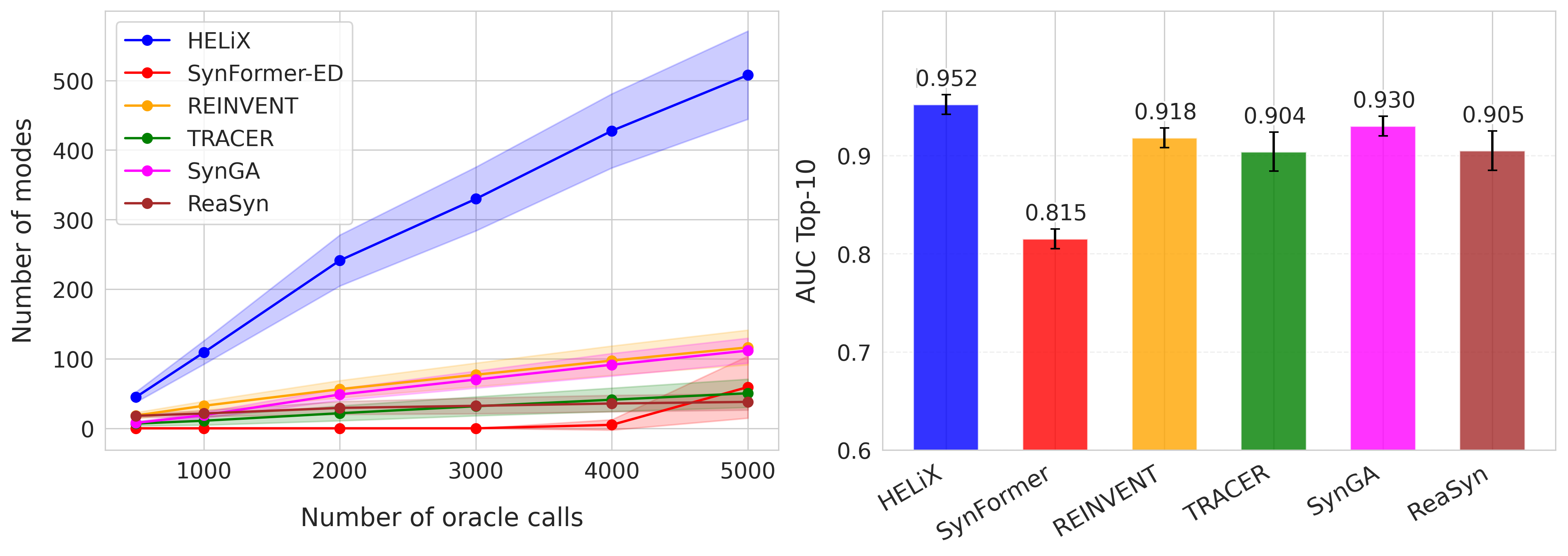}
    \vspace{-15pt}
    \caption{Sample efficiency comparison against baseline methods. Left: number of modes discovered over the first 5k oracle calls. Right: AUC Top-10 comparison against baseline methods evaluated at 10k oracle calls. The shaded regions and error bars indicate $\pm1$ standard deviation across initial hits.}
    \label{fig:sample_efficiency}
    \vspace{-10pt}
\end{figure}

While we previously compared the performance of each method at 10k oracle calls, we are primarily interested in the performance in early stages of optimization. To assess the sample efficiency of HELiX, we plot the number of high-reward modes discovered as a function of oracle calls in Figure~\ref{fig:sample_efficiency} (left), focusing on the first 5k observed molecules. HELiX consistently outperforms all baselines throughout this regime. Additionally, while SynFormer-ED achieves competitive performance at 10k oracle evaluations (Table~\ref{tab:baselines_seh}), it exhibits markedly poorer sample efficiency at fewer oracle calls, failing to discover any high-reward analog modes until approximately 4,000 evaluations.

We further compare methods using the Area Under the Curve (AUC) top-10 metric evaluated at 10k oracle calls \citep{pmo}, which measures how quickly high-quality solutions accumulate among the top-ranked candidates, capturing both solution quality and speed of discovery under a fixed evaluation budget. As shown in Figure~\ref{fig:sample_efficiency} (right), HELiX achieves the highest AUC top-10 among all baselines, indicating superior early-stage performance in terms of exploration---identifying a larger number of diverse, high-reward compounds, as well as exploitation---achieving competitive top-10 mean reward with the strongest baselines (Table~\ref{tab:baselines_seh}).

\subsection{Bayesian Optimization in GFlowNets}
\label{sec:experiments_bo}

We investigate the behavior of BO in GFlowNets by constructing an enumerated library setting in which a small set of reaction templates and building blocks is recursively enumerated to yield a fixed candidate pool of approximately 30k molecules. This allows us to compare direct sampling from the acquisition function to the learned sampling of the GFlowNet policy model operating over the same design space. We study this in a \textit{de novo} setting, optimizing for the JNK3 oracle. Generating from an empty initial state fixes a single search space across all runs, whereas hit-conditioned generation would vary the search space along with the acquisition function. However, we expect the resulting observations about acquisition landscapes to transfer to the hit-conditioned setting, though we do not verify this directly.

\begin{figure}[t]
    \centering
    \includegraphics[width=\linewidth]{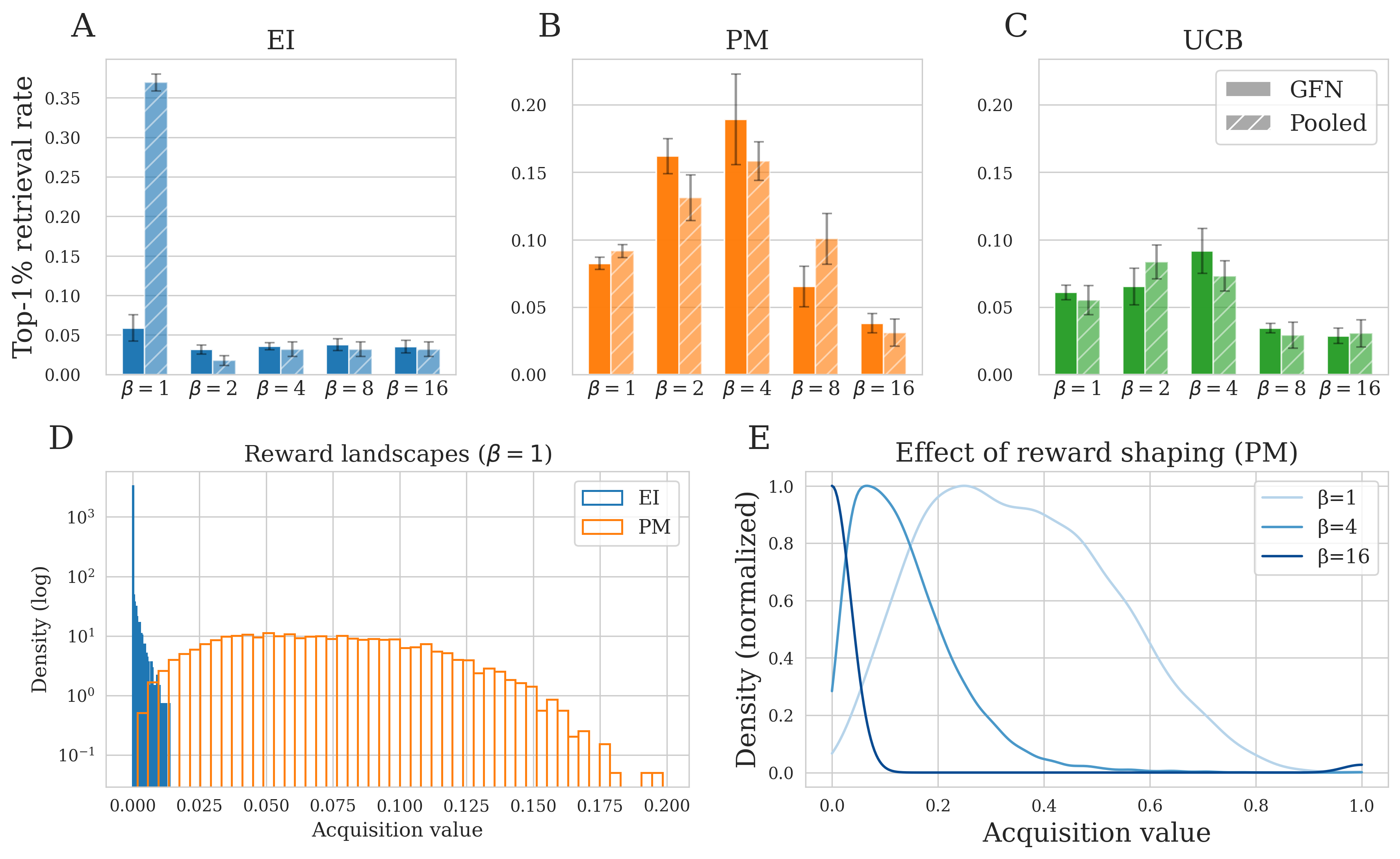}
    \vspace{-15pt}
    \caption{Analysis of acquisition function behavior under proportional sampling. (A--C) Top-1\% retrieval rate for pool-based and GFlowNet samplers under EI, PM, and UCB acquisition functions across reward shaping values $\beta \in \{1, 2, 4, 8, 16\}$. All runs are performed for 20 iterations of BO with a batch size of 50, resulting in 1k observed candidates. Results are averaged over five seeds. (D) Distribution of acquisition values at $\beta = 1$ for EI and PM, showing that EI concentrates near zero while PM exhibits a smoother distribution. Acquisition values are recorded from observations aggregated over all five seeds. (E) Effect of reward shaping on the PM acquisition landscape.}
    \label{fig:bo_figre_1}
    \vspace{-3pt}
\end{figure}

Figure~\ref{fig:bo_figre_1} compares the performance of pool-based and GFlowNet samplers across three acquisition functions---(A) EI, (B) posterior mean (PM), and (C) UCB (defined in Section~\ref{sec:appendix_bo})---and varying degrees of reward shaping. In the pure sampling setting with no reward shaping, EI achieves the highest retrieval rate, which we attribute to its sparse landscape concentrated on high-reward molecules. However, training a GFlowNet with EI results in dramatically lower performance. Alternatively, the performance of a GFlowNet trained with PM or UCB appears to match the pure sampler, suggesting that these acquisition functions provide more suitable reward landscapes for training. Interestingly, applying reward shaping to PM and UCB has a positive effect up to $\beta=4$, with diminishing returns for increased $\beta$. Furthermore, PM appears to consistently outperform UCB, suggesting that the additional exploration in UCB is unnecessary, as sampling provides natural exploration. Moreover, transforming EI does not recover better performance in either setting. We hypothesize that the landscape is already sparse without reward shaping, and transforming it further likely results in diminished batch diversity. Figure~\ref{fig:bo_figre_1} (D) shows the acquisition values of observed candidates for EI and PM with $\beta=1$. We find that EI values are concentrated near zero, demonstrating the sparsity of the acquisition function landscape, whereas PM values exhibit a smoother distribution. Figure~\ref{fig:bo_figre_1} (E) shows how reward shaping affects the landscape of PM. Increasing $\beta$ skews the acquisition values toward zero, concentrating probability mass on a small number of candidates.

\begin{figure}
    \centering
    \includegraphics[width=\linewidth]{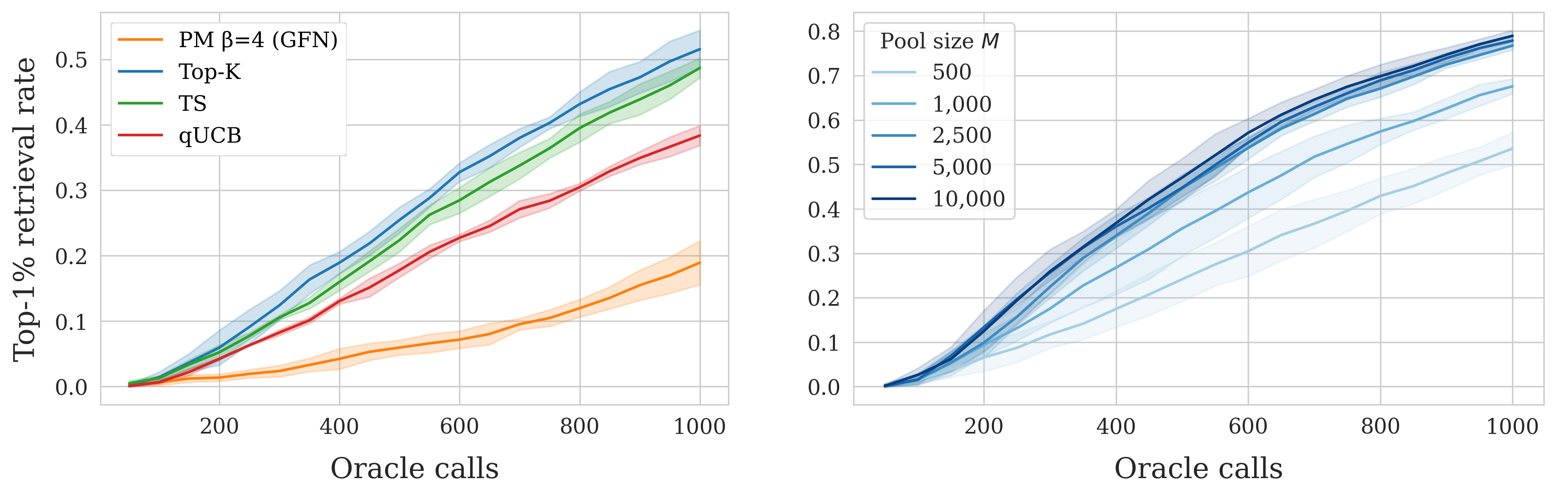}
    \vspace{-10pt}
    \caption{Greedy subsampling for GFlowNet batch selection. (Left) Top-1\% retrieval rate over oracle calls for GFlowNets trained with greedy subsampling methods (Top-$K$, TS, qUCB) against the best reward shaping baseline (PM, $\beta = 4$), with pool size $M = 500$. (Right) Top-1\% retrieval rate over oracle calls for Top-$K$ subsampling with the pool-based sampler. Pool-based results are averaged over three seeds; shaded regions show $\pm 1$ standard deviation. Results for TS and qUCB across pool sizes are shown in Figure~\ref{fig:m_ablation_all}.}
    \label{fig:bo_figure_2}
    \vspace{-10pt}
\end{figure}

Figure~\ref{fig:bo_figure_2} compares GFlowNets trained with three greedy subsampling methods---Top-$K$, Thompson Sampling (TS), and Multi-point UCB (qUCB)---defined in Section~\ref{sec:appendix_bo}, against the best-performing reward shaping baseline (PM with $\beta=4$). With a pool size of $M=500$, all greedy subsampling methods dramatically outperform the GFlowNet with reward shaping: taking the top 50 candidates from the generated pool at each iteration achieves a top-1\% retrieval rate of .5 compared to .2 achieved by the baseline method. This demonstrates an effective alternative to tuning the acquisition landscape: training a GFlowNet on a smooth acquisition function such as PM allows for convergence and broad exploration, while greedy subsampling provides exploitation. To isolate the effect of pool size from GFlowNet training error, the right panel repeats the Top-$K$ experiment with the pool-based sampler while varying $M$. Larger pools improve retrieval throughout the budget, from .54 at $M=500$ to .79 at $M=10$k after 1k oracle calls, with gains saturating beyond $M=2.5$k: since Top-$K$ is applied to a larger sample, it approaches direct maximization of PM over the library. The same trend holds for TS and qUCB (Figure~\ref{fig:m_ablation_all}), and Top-$K$ outperforms both at every pool size, despite having no diversity consideration. We also highlight that, while greedy subsampling with no reward shaping outperforms the best reward shaping baseline in this setting, increasingly large design spaces may benefit from some concentration on high-reward regions, and a successful method would likely combine the two approaches---we leave this exploration to future work.

\begin{wrapfigure}[19]{r}{0.45\textwidth}
    \centering
    \vspace{-10pt}
    \includegraphics[width=\linewidth]{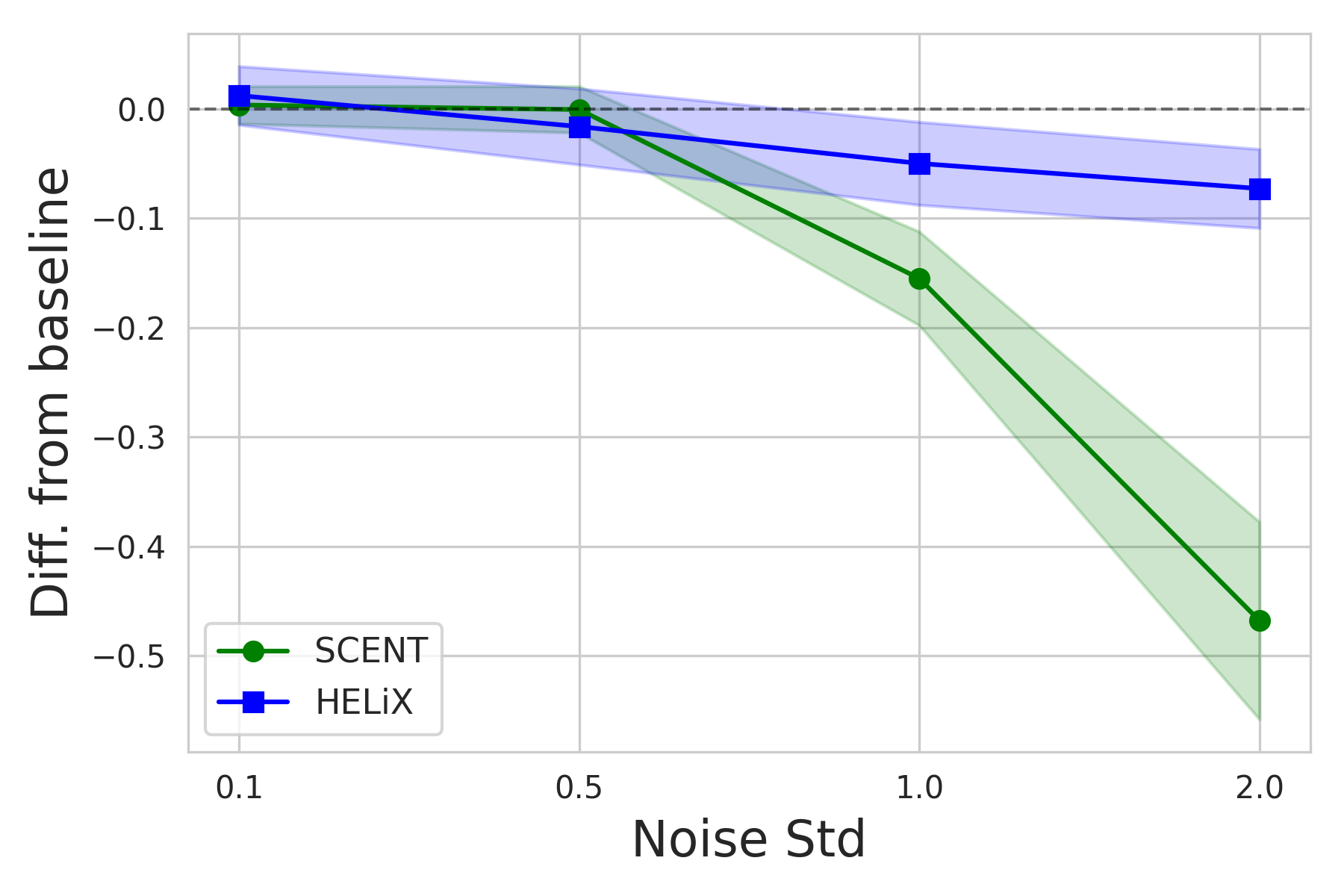}
    \vspace{-10pt}
    \caption{\small Performance degradation of HELiX and SCENT under increasing levels of additive Gaussian noise. Metric is reported as difference in top-10 mean between noisy experiment and baseline (noise-free). Performance is evaluated using the ground-truth (noise-free) proxy.}
    \label{fig:noisy_proxy}
\end{wrapfigure}

\subsection{Robustness to Unreliable Oracles}
\label{sec:noise_robustness}

We further evaluate HELiX when optimizing for unreliable reward signals, simulating the use of imperfect binding affinity proxies such as molecular docking. We hypothesize that searching near known active compounds provides inherent robustness to noise. We test this by comparing HELiX to SCENT under increasing levels of additive Gaussian noise. Metrics are then evaluated using the ground-truth (noise-free) proxy to assess how well each method recovers high-reward compounds despite noisy training signals. 

Figure~\ref{fig:noisy_proxy} shows that HELiX maintains nearly constant performance as the noise increases, while SCENT deteriorates rapidly. Since sEH proxy scores typically range from $0$ to $8.5$, Gaussian noise with standard deviation $\sigma = 2.0$ represents a relatively high noise-to-signal ratio. This demonstrates that anchoring the search space in the local neighborhood of a known active binder provides robust optimization even under highly unreliable proxies.

\subsection{Virtual screening for SETDB1 inhibitors}
\label{sec:experiments_virtual_screening}

Finally, we evaluate HELiX on a real drug discovery target, starting from two experimentally validated SETDB1 binders and optimizing AutoDock Vina docking scores. Unlike the oracles previously considered, docking is comparatively expensive to evaluate and is not a smooth learned function of molecular structure, providing a realistic evaluation of HELiX and the BO-guided training loop. We consider two settings: optimization under a constrained oracle budget, and large-scale screening for a real campaign.

\paragraph{Optimization under a constrained budget.} We compare HELiX with and without BO against REINVENT4, each given a budget of 1k oracle calls, on each of the two hits. HELiX achieves the best docking scores on both hits, with Top-10 scores of $-10.34$ and $-12.93$ kcal/mol, against $-10.01$ and $-10.13$ without BO and $-9.47$ and $-10.68$ for REINVENT4 (Table~\ref{tab:setdb1_bo}). BO appears to contribute meaningfully to the sample efficiency. This indicates that the benefits of BO-guided training extend beyond smooth learned proxies. A remaining concern is that high docking scores may reflect reward hacking rather than improved predicted binding. To test this, we rescore the top candidates from each method with GNINA \citep{mcnutt2021gnina}, an independent scoring function never used during optimization. We find that HELiX's molecules transfer to higher scores on this held-out oracle, reaching predicted binding affinities of 6.29 and 6.23 compared to 5.91 and 5.51 for REINVENT4 (Appendix~\ref{sec:heldout_oracle}).

\begin{figure}[h]
    \centering
    \vspace{-5pt}
    \includegraphics[width=.86\linewidth]{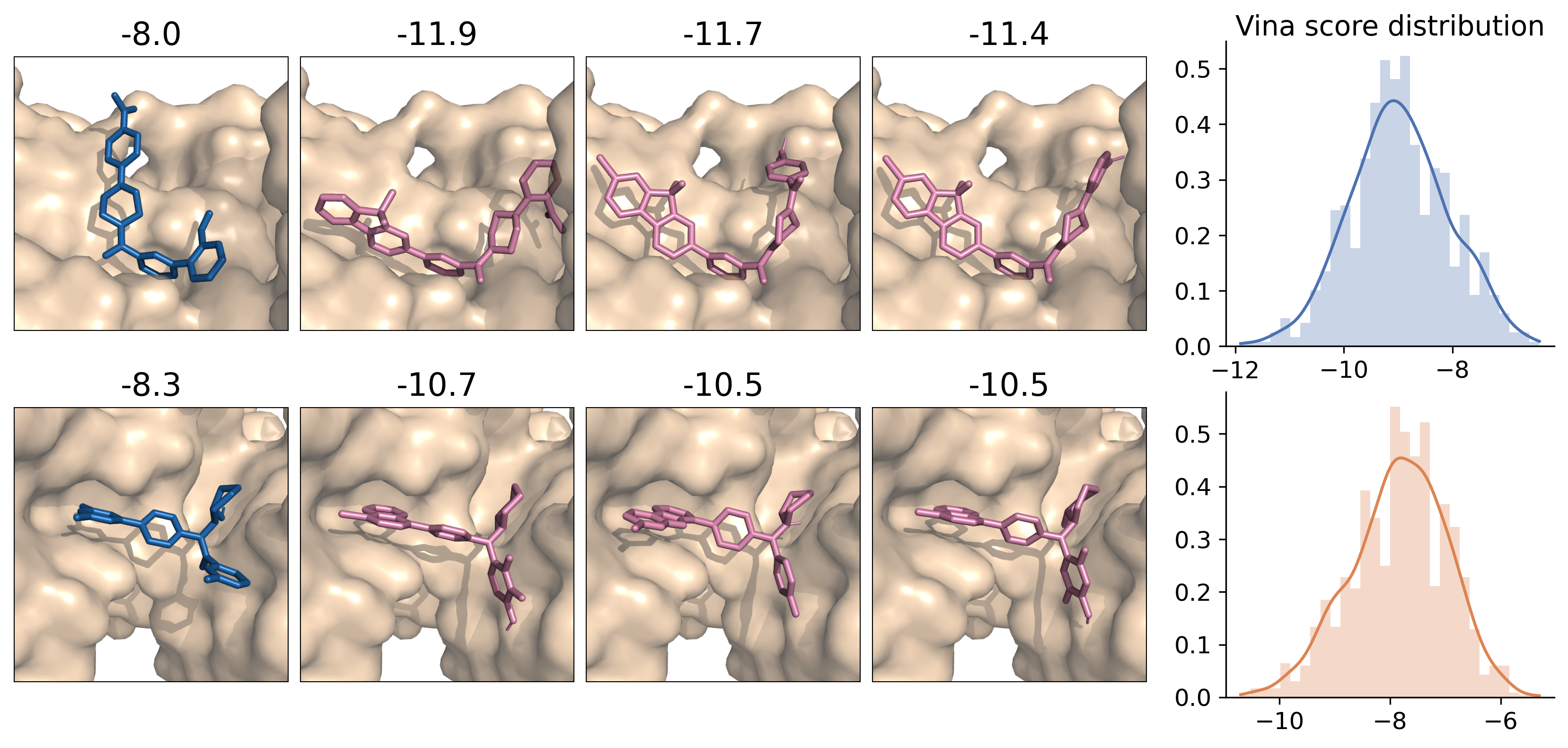}
    \vspace{-10pt}
    \caption{Binding poses of the two SETDB1 hits (blue) and top-3 analogs generated by HELiX (pink), docked to the 7CJT conformation with AutoDock Vina scores (kcal/mol) displayed above. Right: distribution of Vina scores across all generated candidates for each hit.}
    \label{fig:qualitative_binders}
    \vspace{-10pt}
\end{figure}

\paragraph{Screening at scale.}
Separately, we used HELiX to generate candidates for downstream experimental screening. Three protein conformations were targeted (PDB: 7CJT, 8G5E, 8UWP), yielding 6 settings (2 hits $\times$ 3 conformations). For each setting, HELiX was used in virtual screening by optimizing AutoDock Vina scores directly without the use of BO since sample efficiency was not a primary concern. Optimization was performed across three hyperparameter settings in parallel (reward $\beta \in \{2, 4, 6\}$). The generated molecules were aggregated and de-duplicated, resulting in roughly 30k candidates for each hit and conformation. Candidates were then filtered to have molecular weight (MW) between 300-550 Da, and QED $\geq$ 0.35. The top 100 molecules by docking score were retained for each setting, yielding a concatenated set of 600 total candidates to be proposed for downstream analysis. The top 3 candidates generated from each hit targeting the 7CJT conformation are shown in Figure~\ref{fig:qualitative_binders}; we show only 7CJT for brevity. Across retained candidates, many docking scores improve substantially over the initial hits (Figure~\ref{fig:qualitative_binders}, right), with the top analogs reaching $-11.9$ and $-10.7$ kcal/mol against $-8.0$ and $-8.3$ for the two hits. We furthermore assess the physical validity of these poses in Appendix~\ref{sec:posebusters}.

\section{Conclusion}
\label{sec:conclusion}

We present HELiX, a template-based GFlowNet framework for local hit expansion in synthesizable chemical space. HELiX achieves state-of-the-art performance in identifying diverse, high-reward analogs of known binders under limited oracle budgets. We further demonstrate robustness to noisy oracle evaluations, an inherent benefit of anchoring search to the local neighborhood of known actives, as well as applicability to a realistic virtual screening campaign. A key contribution of this work is a systematic study of the failure modes that arise when training GFlowNets with standard BO acquisition functions. We show that sparse acquisition landscapes, such as those produced by EI, impede GFlowNet training convergence. As an alternative, we introduce greedy acquisition subsampling which outperforms reward shaping baselines.

\paragraph{Limitations.} Our BO analysis is conducted in a small enumerated library setting; behavior in the full HELiX generation space warrants further study. With the exception of the SETDB1 case study, oracle evaluations use pre-trained proxy models rather than experimental assays, and performance is contingent on hits being decomposable under the reaction template library. We also optimize a single objective throughout: as shown in Section~\ref{sec:synth_metrics}, properties outside the reward degrade relative to the initial hit, and a practical campaign would likely require a multi-objective formulation incorporating drug-likeness or selectivity terms. Finally, we show a proof-of-concept combining HELiX with a projection step to address hits which are not decomposable; however, the performance of this method depends on the ability of the projection method to find close analogs of the non-synthesizable reference molecule. We leave these challenges to future work.

\section*{Acknowledgements and Disclosures}

\paragraph{Funding.} This research was supported by the Gates Foundation (grant number INV-092957); Mitacs, through the Mitacs Accelerate program; the Acceleration Consortium from the Canada First Research Excellence Fund (grant number CFREF-2022-00042); and the Canadian Institutes for Health Research (grant number FDN-167277). W.V. is supported by a Queen Elizabeth II Graduate Scholarship in Science and Technology. Computational resources were provided by the Digital Research Alliance of Canada. 

\paragraph{Competing interests.} The authors declare no competing interests.

\paragraph{Dual Use.} HELiX is designed to accelerate therapeutic drug discovery; however, there exists a potential for misuse---for instance, applying similar methods to design harmful compounds. Nevertheless, we feel that the potential benefits outweigh the harm.

\paragraph{Code Availability.} Code is available at \href{https://github.com/koziarskilab/helix}{https://github.com/koziarskilab/helix}.

\bibliographystyle{plainnat}
\bibliography{references}

\begin{thebibliography}{56}
\providecommand{\natexlab}[1]{#1}
\providecommand{\url}[1]{\texttt{#1}}
\expandafter\ifx\csname urlstyle\endcsname\relax
  \providecommand{\doi}[1]{doi: #1}\else
  \providecommand{\doi}{doi: \begingroup \urlstyle{rm}\Url}\fi

\bibitem[AI4Science et~al.(2023)AI4Science, Hernandez-Garcia, Duval, Volokhova,
  Bengio, Sharma, Carrier, Benabed, Koziarski, and
  Schmidt]{ai4science2023crystal}
Mila AI4Science, Alex Hernandez-Garcia, Alexandre Duval, Alexandra Volokhova,
  Yoshua Bengio, Divya Sharma, Pierre~Luc Carrier, Yasmine Benabed, Micha{\l}
  Koziarski, and Victor Schmidt.
\newblock Crystal-{GFN}: sampling crystals with desirable properties and
  constraints.
\newblock \emph{arXiv preprint arXiv:2310.04925}, 2023.

\bibitem[Balandat et~al.(2020)Balandat, Karrer, Jiang, Daulton, Letham, Wilson,
  and Bakshy]{balandat_botorch_2020}
Maximilian Balandat, Brian Karrer, Daniel~R. Jiang, Samuel Daulton, Benjamin
  Letham, Andrew~Gordon Wilson, and Eytan Bakshy.
\newblock {BoTorch}: {A} {Framework} for {Efficient} {Monte}-{Carlo} {Bayesian}
  {Optimization}, December 2020.
\newblock URL \url{http://arxiv.org/abs/1910.06403}.
\newblock arXiv:1910.06403 [cs].

\bibitem[Bemis and Murcko(1996)]{bemis1996properties}
Guy~W. Bemis and Mark~A. Murcko.
\newblock The properties of known drugs. 1. molecular frameworks.
\newblock \emph{Journal of Medicinal Chemistry}, 39\penalty0 (15):\penalty0
  2887--2893, 1996.
\newblock \doi{10.1021/jm9602928}.

\bibitem[Bengio et~al.(2021)Bengio, Jain, Korablyov, Precup, and
  Bengio]{bengio2021}
Emmanuel Bengio, Moksh Jain, Maksym Korablyov, Doina Precup, and Yoshua Bengio.
\newblock Flow network based generative models for non-iterative diverse
  candidate generation.
\newblock In \emph{Advances in Neural Information Processing Systems 34
  (NeurIPS 2021)}, 2021.

\bibitem[Bengio et~al.(2023)Bengio, Lahlou, Deleu, Hu, Tiwari, and
  Bengio]{bengio2023}
Yoshua Bengio, Salem Lahlou, Tristan Deleu, Edward~J. Hu, Mo~Tiwari, and
  Emmanuel Bengio.
\newblock {GFlowNet} foundations.
\newblock \emph{Journal of Machine Learning Research}, 24\penalty0
  (210):\penalty0 1--55, 2023.

\bibitem[Bradshaw et~al.(2019)Bradshaw, Paige, Kusner, Segler, and
  Hern{\'a}ndez-Lobato]{bradshaw2019model}
John Bradshaw, Brooks Paige, Matt~J Kusner, Marwin Segler, and Jos{\'e}~Miguel
  Hern{\'a}ndez-Lobato.
\newblock A model to search for synthesizable molecules.
\newblock \emph{Advances in Neural Information Processing Systems}, 32, 2019.

\bibitem[Bradshaw et~al.(2020)Bradshaw, Paige, Kusner, Segler, and
  Hern{\'a}ndez-Lobato]{bradshaw2020barking}
John Bradshaw, Brooks Paige, Matt~J Kusner, Marwin Segler, and Jos{\'e}~Miguel
  Hern{\'a}ndez-Lobato.
\newblock Barking up the right tree: an approach to search over molecule
  synthesis {DAG}s.
\newblock \emph{Advances in Neural Information Processing systems},
  33:\penalty0 6852--6866, 2020.

\bibitem[Buttenschoen et~al.(2024)Buttenschoen, Morris, and
  Deane]{buttenschoen2024posebusters}
Martin Buttenschoen, Garrett~M. Morris, and Charlotte~M. Deane.
\newblock {{PoseBusters}}: {{AI-based}} docking methods fail to generate
  physically valid poses or generalise to novel sequences.
\newblock \emph{Chemical Science}, 15:\penalty0 3130--3139, 2024.
\newblock \doi{10.1039/D3SC04185A}.
\newblock URL \url{http://dx.doi.org/10.1039/D3SC04185A}.

\bibitem[Cretu et~al.(2025)Cretu, Harris, Igashov, Schneuing, Segler, Correia,
  Roy, Bengio, and Lio]{cretu2025synflownet}
Miruna Cretu, Charles Harris, Ilia Igashov, Arne Schneuing, Marwin Segler,
  Bruno Correia, Julien Roy, Emmanuel Bengio, and Pietro Lio.
\newblock Synflownet: Design of diverse and novel molecules with synthesis
  constraints.
\newblock In \emph{The Thirteenth International Conference on Learning
  Representations}, 2025.
\newblock URL \url{https://openreview.net/forum?id=uvHmnahyp1}.

\bibitem[Degen et~al.(2008)Degen, Wegscheid-Gerlach, Zaliani, and
  Rarey]{degen2008brics}
J{\"o}rg Degen, Christof Wegscheid-Gerlach, Andrea Zaliani, and Matthias Rarey.
\newblock On the art of compiling and using `drug-like' chemical fragment
  spaces.
\newblock \emph{ChemMedChem}, 3\penalty0 (10):\penalty0 1503--1507, 2008.
\newblock \doi{10.1002/cmdc.200800178}.

\bibitem[Eberhardt et~al.(2021)Eberhardt, Santos-Martins, Tillack, and
  Forli]{vina2}
Jerome Eberhardt, Diogo Santos-Martins, Andreas~F. Tillack, and Stefano Forli.
\newblock Autodock vina 1.2.0: New docking methods, expanded force field, and
  python bindings.
\newblock \emph{Journal of Chemical Information and Modeling}, 2021.
\newblock \doi{10.1021/acs.jcim.1c00203}.

\bibitem[Ertl and Schuffenhauer(2009)]{ertl2009estimation}
Peter Ertl and Ansgar Schuffenhauer.
\newblock Estimation of synthetic accessibility score of drug-like molecules
  based on molecular complexity and fragment contributions.
\newblock \emph{Journal of cheminformatics}, 1:\penalty0 1--11, 2009.

\bibitem[Gai{\'n}ski et~al.(2025{\natexlab{a}})Gai{\'n}ski, Boussif, Rekesh,
  Shevchuk, Parviz, Tyers, Batey, and Koziarski]{gainski2025scalable}
Piotr Gai{\'n}ski, Oussama Boussif, Andrei Rekesh, Dmytro Shevchuk, Ali Parviz,
  Mike Tyers, Robert~A. Batey, and Micha{\l} Koziarski.
\newblock Scalable and cost-efficient de novo template-based molecular
  generation.
\newblock In \emph{The Thirty-ninth Annual Conference on Neural Information
  Processing Systems}, 2025{\natexlab{a}}.
\newblock URL \url{https://openreview.net/forum?id=zssWxiiJZ1}.

\bibitem[Gai{\'n}ski et~al.(2025{\natexlab{b}})Gai{\'n}ski, Koziarski, Maziarz,
  Segler, Tabor, and {\'S}mieja]{gainski2025diverse}
Piotr Gai{\'n}ski, Micha{\l} Koziarski, Krzysztof Maziarz, Marwin Segler, Jacek
  Tabor, and Marek {\'S}mieja.
\newblock Diverse and feasible retrosynthesis using {GFlowNets}.
\newblock \emph{Information Sciences}, 714:\penalty0 122194,
  2025{\natexlab{b}}.

\bibitem[Gao et~al.(2022{\natexlab{a}})Gao, Fu, Sun, and Coley]{pmo}
Wenhao Gao, Tianfan Fu, Jimeng Sun, and Connor~W. Coley.
\newblock Sample efficiency matters: A benchmark for practical molecular
  optimization, 2022{\natexlab{a}}.

\bibitem[Gao et~al.(2022{\natexlab{b}})Gao, Mercado, and
  Coley]{gao2021amortized}
Wenhao Gao, Roc{\'\i}o Mercado, and Connor~W Coley.
\newblock Amortized tree generation for bottom-up synthesis planning and
  synthesizable molecular design.
\newblock In \emph{The Tenth International Conference on Learning
  Representations}, 2022{\natexlab{b}}.

\bibitem[Gao et~al.(2024)Gao, Luo, and Coley]{gao2024synformer}
Wenhao Gao, Shitong Luo, and Connor~W. Coley.
\newblock Generative ai for navigating synthesizable chemical space.
\newblock \emph{Proceedings of the National Academy of Sciences}, 121:\penalty0
  e2415665122, 2024.

\bibitem[Garnett(2023)]{garnett_bayesoptbook_2023}
Roman Garnett.
\newblock \emph{{Bayesian Optimization}}.
\newblock Cambridge University Press, 2023.

\bibitem[Genheden et~al.(2020)Genheden, Thakkar, Chadimov{\'a}, Reymond,
  Engkvist, and Bjerrum]{genheden2020aizynthfinder}
Samuel Genheden, Amol Thakkar, Veronika Chadimov{\'a}, Jean-Louis Reymond, Ola
  Engkvist, and Esben Bjerrum.
\newblock {AiZynthFinder}: a fast, robust and flexible open-source software for
  retrosynthetic planning.
\newblock \emph{Journal of cheminformatics}, 12\penalty0 (1):\penalty0 70,
  2020.

\bibitem[Ghari et~al.(2024)Ghari, Tseng, Eraslan, Lopez, Biancalani, Scalia,
  and Hajiramezanali]{ghari2024}
Pouya~M. Ghari, Alex~M. Tseng, G\"{o}kcen Eraslan, Romain Lopez, Tommaso
  Biancalani, Gabriele Scalia, and Ehsan Hajiramezanali.
\newblock Gflownet assisted biological sequence editing.
\newblock In \emph{Advances in Neural Information Processing Systems 37
  (NeurIPS 2024)}, 2024.

\bibitem[G{\'o}mez-Bombarelli et~al.(2018)G{\'o}mez-Bombarelli, Wei, Duvenaud,
  Hern{\'a}ndez-Lobato, S{\'a}nchez-Lengeling, Sheberla, Aguilera-Iparraguirre,
  Hirzel, Adams, and Aspuru-Guzik]{GomezBombarelli2018}
Rafael G{\'o}mez-Bombarelli, Jennifer~N Wei, David Duvenaud, Jos{\'e}~Miguel
  Hern{\'a}ndez-Lobato, Benjam{\'\i}n S{\'a}nchez-Lengeling, Dennis Sheberla,
  Jorge Aguilera-Iparraguirre, Timothy~D Hirzel, Ryan~P Adams, and Al{\'a}n
  Aspuru-Guzik.
\newblock Automatic chemical design using a data-driven continuous
  representation of molecules.
\newblock \emph{ACS central science}, 4\penalty0 (2):\penalty0 268--276, 2018.

\bibitem[Gottipati et~al.(2020)Gottipati, Sattarov, Niu, Pathak, Wei, Liu,
  Blackburn, Thomas, Coley, Tang, et~al.]{gottipati2020learning}
Sai~Krishna Gottipati, Boris Sattarov, Sufeng Niu, Yashaswi Pathak, Haoran Wei,
  Shengchao Liu, Simon Blackburn, Karam Thomas, Connor Coley, Jian Tang, et~al.
\newblock Learning to navigate the synthetically accessible chemical space
  using reinforcement learning.
\newblock In \emph{International conference on machine learning}, pages
  3668--3679. PMLR, 2020.

\bibitem[Guo and Schwaller(2025)]{guo2025directly}
Jeff Guo and Philippe Schwaller.
\newblock Directly optimizing for synthesizability in generative molecular
  design using retrosynthesis models.
\newblock \emph{Chemical Science}, 16\penalty0 (16):\penalty0 6943--6956, 2025.

\bibitem[Hernandez-Garcia et~al.(2024)Hernandez-Garcia, Saxena, Jain, Liu, and
  Bengio]{hernandez-garcia_mfal_2024}
Alex Hernandez-Garcia, Nikita Saxena, Moksh Jain, Cheng-Hao Liu, and Yoshua
  Bengio.
\newblock {MFAL} w/ {GFlowNets}, 2024.

\bibitem[Horwood and Noutahi(2020)]{horwood2020molecular}
Julien Horwood and Emmanuel Noutahi.
\newblock Molecular design in synthetically accessible chemical space via deep
  reinforcement learning.
\newblock \emph{ACS omega}, 5\penalty0 (51):\penalty0 32984--32994, 2020.

\bibitem[Huang et~al.(2021)Huang, Fu, Gao, Zhao, Roohani, Leskovec, Coley,
  Xiao, Sun, and Zitnik]{tdc}
Kexin Huang, Tianfan Fu, Wenhao Gao, Yue Zhao, Yusuf Roohani, Jure Leskovec,
  Connor~W Coley, Cao Xiao, Jimeng Sun, and Marinka Zitnik.
\newblock Therapeutics data commons: Machine learning datasets and tasks for
  drug discovery and development.
\newblock \emph{arXiv preprint arXiv:2102.09548}, 2021.

\bibitem[Jain et~al.(2023)Jain, Bengio, Garcia, Rector-Brooks, Dossou, Ekbote,
  Fu, Zhang, Kilgour, Zhang, Simine, Das, and Bengio]{jain_biological_2023}
Moksh Jain, Emmanuel Bengio, Alex-Hernandez Garcia, Jarrid Rector-Brooks,
  Bonaventure F.~P. Dossou, Chanakya Ekbote, Jie Fu, Tianyu Zhang, Micheal
  Kilgour, Dinghuai Zhang, Lena Simine, Payel Das, and Yoshua Bengio.
\newblock Biological {Sequence} {Design} with {GFlowNets}, 2023.

\bibitem[Jocys et~al.(2024)Jocys, Willems, and Farrahi]{jocys2024synthformer}
Zygimantas Jocys, Henriette~MG Willems, and Katayoun Farrahi.
\newblock Synth{F}ormer: Equivariant pharmacophore-based generation of
  molecules for ligand-based drug design.
\newblock \emph{arXiv preprint arXiv:2410.02718}, 2024.

\bibitem[Kim et~al.(2024)Kim, Kim, Choi, and Park]{kim2024}
Hyeonah Kim, Minsu Kim, Sanghyeok Choi, and Jinkyoo Park.
\newblock Genetic-guided gflownets for sample efficient molecular optimization.
\newblock In \emph{Advances in Neural Information Processing Systems 37
  (NeurIPS 2024)}, 2024.

\bibitem[Kim et~al.(2026)Kim, Kim, Roget, Biton, Vaillancourt, Brun, Bengio,
  and Hernandez-Garcia]{kim2026synthesizablemoleculargenerationsoftconstrained}
Hyeonah Kim, Minsu Kim, Celine Roget, Dionessa Biton, Louis Vaillancourt,
  Yves~V. Brun, Yoshua Bengio, and Alex Hernandez-Garcia.
\newblock Synthesizable molecular generation via soft-constrained gflownets
  with rich chemical priors, 2026.
\newblock URL \url{https://arxiv.org/abs/2602.04119}.

\bibitem[Korablyov et~al.(2024)Korablyov, Liu, Jain, van~der Sloot, Jolicoeur,
  Ruediger, Nica, Bengio, Lapchevskyi, St-Cyr, et~al.]{korablyov2024generative}
Maksym Korablyov, Cheng-Hao Liu, Moksh Jain, Almer~M van~der Sloot, Eric
  Jolicoeur, Edward Ruediger, Andrei~Cristian Nica, Emmanuel Bengio, Kostiantyn
  Lapchevskyi, Daniel St-Cyr, et~al.
\newblock Generative active learning for the search of small-molecule protein
  binders.
\newblock \emph{arXiv preprint arXiv:2405.01616}, 2024.

\bibitem[Koziarski et~al.(2024)Koziarski, Rekesh, Shevchuk, van~der Sloot,
  Gai\'{n}ski, Bengio, Liu, Tyers, and Batey]{koziarski2024}
Micha\l{} Koziarski, Andrei Rekesh, Dmytro Shevchuk, Almer van~der Sloot, Piotr
  Gai\'{n}ski, Yoshua Bengio, Cheng-Hao Liu, Mike Tyers, and Robert~A. Batey.
\newblock Rgfn: Synthesizable molecular generation using gflownets.
\newblock In \emph{Advances in Neural Information Processing Systems 37
  (NeurIPS 2024)}, 2024.

\bibitem[Lee et~al.(2026)Lee, Kreis, Veccham, Liu, Reidenbach, Paliwal, Nie,
  and Vahdat]{lee_exploring_2026}
Seul Lee, Karsten Kreis, Srimukh~Prasad Veccham, Meng Liu, Danny Reidenbach,
  Saee Paliwal, Weili Nie, and Arash Vahdat.
\newblock Exploring {Synthesizable} {Chemical} {Space} with {Iterative}
  {Pathway} {Refinements}, 2026.

\bibitem[Liu et~al.(2022)Liu, Korablyov, Jastrzebski, W{\l}odarczyk-Pruszynski,
  Bengio, and Segler]{liu2022retrognn}
Cheng-Hao Liu, Maksym Korablyov, Stanis{\l}aw Jastrzebski, Pawe{\l}
  W{\l}odarczyk-Pruszynski, Yoshua Bengio, and Marwin Segler.
\newblock {RetroGNN}: fast estimation of synthesizability for virtual screening
  and de novo design by learning from slow retrosynthesis software.
\newblock \emph{Journal of Chemical Information and Modeling}, 62\penalty0
  (10):\penalty0 2293--2300, 2022.

\bibitem[Lo et~al.(2025)Lo, Coley, and Matusik]{lo2025synga}
Alston Lo, Connor~W Coley, and Wojciech Matusik.
\newblock A genetic algorithm for navigating synthesizable molecular spaces.
\newblock \emph{arXiv preprint arXiv:2509.20719}, 2025.

\bibitem[Loeffler et~al.(2024)Loeffler, He, Tibo, Janet, Voronov, Mervin, and
  Engkvist]{loeffler2024reinvent}
Hannes~H. Loeffler, Jiazhen He, Alessandro Tibo, Jon~Paul Janet, Alexey
  Voronov, Lewis~H. Mervin, and Ola Engkvist.
\newblock {REINVENT} 4: Modern ai--driven generative molecule design.
\newblock \emph{Journal of Cheminformatics}, 16\penalty0 (1):\penalty0 20,
  2024.

\bibitem[Luo et~al.(2024)Luo, Gao, Wu, Peng, Coley, and Ma]{luo2024projecting}
Shitong Luo, Wenhao Gao, Zuofan Wu, Jian Peng, Connor~W Coley, and Jianzhu Ma.
\newblock Projecting molecules into synthesizable chemical spaces.
\newblock In \emph{International Conference on Machine Learning}, pages
  33289--33304. PMLR, 2024.

\bibitem[Malkin et~al.(2022)Malkin, Jain, Bengio, Sun, and Bengio]{malkin2022}
Nikolay Malkin, Moksh Jain, Emmanuel Bengio, Chen Sun, and Yoshua Bengio.
\newblock Trajectory balance: Improved credit assignment in gflownets.
\newblock In \emph{Advances in Neural Information Processing Systems 35
  (NeurIPS 2022)}, pages 5955--5967, 2022.

\bibitem[McNutt et~al.(2021)McNutt, Francoeur, Aggarwal, Masuda, Meli, Ragoza,
  Sunseri, and Koes]{mcnutt2021gnina}
Andrew~T. McNutt, Paul Francoeur, Rishal Aggarwal, Tomohide Masuda, Rocco Meli,
  Matthew Ragoza, Jocelyn Sunseri, and David~Ryan Koes.
\newblock {GNINA} 1.0: molecular docking with deep learning.
\newblock \emph{Journal of Cheminformatics}, 13\penalty0 (1):\penalty0 43,
  2021.
\newblock \doi{10.1186/s13321-021-00522-2}.

\bibitem[Nakamura et~al.(2025)Nakamura, Yasuo, and
  Sekijima]{nakamura2025tracer}
Shogo Nakamura, Nobuaki Yasuo, and Masakazu Sekijima.
\newblock Molecular optimization using a conditional transformer for
  reaction-aware compound exploration with reinforcement learning.
\newblock \emph{Communications Chemistry}, 8:\penalty0 40, 2025.

\bibitem[Olivecrona et~al.(2017)Olivecrona, Blaschke, Engkvist, and
  Chen]{Olivecrona2017}
Marcus Olivecrona, Thomas Blaschke, Ola Engkvist, and Hongming Chen.
\newblock Molecular de-novo design through deep reinforcement learning.
\newblock \emph{Journal of cheminformatics}, 9\penalty0 (1):\penalty0 48, 2017.

\bibitem[Rekesh et~al.(2025)Rekesh, Cretu, Shevchuk, Somnath, Li{\`o}, Batey,
  Tyers, Koziarski, and Liu]{rekesh2025syncogen}
Andrei Rekesh, Miruna Cretu, Dmytro Shevchuk, Vignesh~Ram Somnath, Pietro
  Li{\`o}, Robert~A Batey, Mike Tyers, Micha{\l} Koziarski, and Cheng-Hao Liu.
\newblock {SynCoGen}: Synthesizable {3D} molecule generation via joint reaction
  and coordinate modeling.
\newblock \emph{arXiv preprint arXiv:2507.11818}, 2025.

\bibitem[Seo et~al.(2023)Seo, Lim, and Kim]{seo2023molecular}
Seonghwan Seo, Jaechang Lim, and Woo~Youn Kim.
\newblock Molecular generative model via retrosynthetically prepared chemical
  building block assembly.
\newblock \emph{Advanced Science}, 10\penalty0 (8):\penalty0 2206674, 2023.

\bibitem[Seo et~al.(2025)Seo, Kim, Shen, Ester, Park, Ahn, and
  Kim]{seo2025generative}
Seonghwan Seo, Minsu Kim, Tony Shen, Martin Ester, Jinkyoo Park, Sungsoo Ahn,
  and Woo~Youn Kim.
\newblock Generative flows on synthetic pathway for drug design.
\newblock In \emph{The Thirteenth International Conference on Learning
  Representations}, 2025.
\newblock URL \url{https://openreview.net/forum?id=pB1XSj2y4X}.

\bibitem[Shahriari et~al.(2015)Shahriari, Swersky, Wang, Adams, and
  De~Freitas]{shahriari2015taking}
Bobak Shahriari, Kevin Swersky, Ziyu Wang, Ryan~P Adams, and Nando De~Freitas.
\newblock Taking the human out of the loop: A review of bayesian optimization.
\newblock \emph{Proceedings of the IEEE}, 104\penalty0 (1):\penalty0 148--175,
  2015.

\bibitem[Shen et~al.(2024)Shen, Seo, Lee, Pandey, Smith, Cherkasov, Kim, and
  Ester]{shen2024}
Tony Shen, Seonghwan Seo, Grayson Lee, Mohit Pandey, Jason~R. Smith, Artem
  Cherkasov, Woo~Youn Kim, and Martin Ester.
\newblock {TacoGFN}: Target conditioned {GFlowNet} for structure-based drug
  design.
\newblock \emph{Transactions on Machine Learning Research}, 2024.

\bibitem[St{\aa}hl et~al.(2019)St{\aa}hl, Falkman, Karlsson, Mathiason, and
  Bostrom]{Stahl2019}
Niclas St{\aa}hl, Goran Falkman, Alexander Karlsson, Gunnar Mathiason, and
  Jonas Bostrom.
\newblock Deep reinforcement learning for multiparameter optimization in de
  novo drug design.
\newblock \emph{Journal of chemical information and modeling}, 59\penalty0
  (7):\penalty0 3166--3176, 2019.

\bibitem[Tom et~al.(2024)Tom, Schmid, Baird, Cao, Darvish, Hao, Lo,
  Pablo-Garc{\'\i}a, Rajaonson, Skreta, et~al.]{tom2024self}
Gary Tom, Stefan~P Schmid, Sterling~G Baird, Yang Cao, Kourosh Darvish, Han
  Hao, Stanley Lo, Sergio Pablo-Garc{\'\i}a, Ella~M Rajaonson, Marta Skreta,
  et~al.
\newblock Self-driving laboratories for chemistry and materials science.
\newblock \emph{Chemical Reviews}, 124\penalty0 (16):\penalty0 9633--9732,
  2024.

\bibitem[Tong et~al.(2021)Tong, Liu, Tan, Li, Jiang, Xiong, Xu, Jiang, Qiao,
  and Zheng]{tong2021generative}
Xiaochu Tong, Xiaohong Liu, Xiaoqin Tan, Xutong Li, Jiaxin Jiang, Zhaoping
  Xiong, Tingyang Xu, Hualiang Jiang, Nan Qiao, and Mingyue Zheng.
\newblock Generative models for de novo drug design.
\newblock \emph{Journal of Medicinal Chemistry}, 64\penalty0 (19):\penalty0
  14011--14027, 2021.

\bibitem[Tripp and Hern{\'a}ndez-Lobato(2023)]{tripp2023genetic}
Austin Tripp and Jos{\'e}~Miguel Hern{\'a}ndez-Lobato.
\newblock Genetic algorithms are strong baselines for molecule generation.
\newblock \emph{arXiv preprint arXiv:2310.09267}, 2023.

\bibitem[Tversky(1977)]{tversky1977features}
Amos Tversky.
\newblock Features of similarity.
\newblock \emph{Psychological Review}, 84\penalty0 (4):\penalty0 327--352,
  1977.
\newblock \doi{10.1037/0033-295X.84.4.327}.

\bibitem[van~den Broek et~al.(2025)van~den Broek, Patel, van Westen, Jespers,
  and Sherman]{van2025search}
Remco~L van~den Broek, Shivam Patel, Gerard~JP van Westen, Willem Jespers, and
  Woody Sherman.
\newblock In search of beautiful molecules: a perspective on generative
  modeling for drug design.
\newblock \emph{Journal of chemical information and modeling}, 65\penalty0
  (18):\penalty0 9383--9397, 2025.

\bibitem[Volokhova et~al.(2024)Volokhova, Koziarski, Hern{\'a}ndez-Garc{\'\i}a,
  Liu, Miret, Lemos, Thiede, Yan, Aspuru-Guzik, and
  Bengio]{volokhova2024towards}
Alexandra Volokhova, Micha{\l} Koziarski, Alex Hern{\'a}ndez-Garc{\'\i}a,
  Cheng-Hao Liu, Santiago Miret, Pablo Lemos, Luca Thiede, Zichao Yan, Al{\'a}n
  Aspuru-Guzik, and Yoshua Bengio.
\newblock Towards equilibrium molecular conformation generation with
  {GFlowNets}.
\newblock \emph{Digital Discovery}, 3\penalty0 (5):\penalty0 1038--1047, 2024.

\bibitem[Williams(1992)]{williams1992simple}
Ronald~J Williams.
\newblock Simple statistical gradient-following algorithms for connectionist
  reinforcement learning.
\newblock \emph{Machine learning}, 8\penalty0 (3):\penalty0 229--256, 1992.

\bibitem[You et~al.(2018)You, Liu, Ying, Pande, and Leskovec]{You2018}
Jiaxuan You, Bowen Liu, Zhitao Ying, Vijay Pande, and Jure Leskovec.
\newblock Graph convolutional policy network for goal-directed molecular graph
  generation.
\newblock \emph{Advances in neural information processing systems}, 31, 2018.

\bibitem[Zhu et~al.(2026)Zhu, Li, Chen, Zhao, Wang, Li, Liu, Kong, Zhang, Liu,
  Hou, and Hsieh]{zhu_syngfn_2026}
Yuchen Zhu, Shuwang Li, Jihong Chen, Donghai Zhao, Xiaorui Wang, Yitong Li,
  Yifei Liu, Yue Kong, Beichen Zhang, Chang Liu, Tingjun Hou, and Chang-Yu
  Hsieh.
\newblock {SynGFN}: learning across chemical space with generative flow-based
  molecular discovery.
\newblock \emph{Nature Computational Science}, 6\penalty0 (1):\penalty0 29--38,
  January 2026.
\newblock ISSN 2662-8457.
\newblock \doi{10.1038/s43588-025-00902-w}.
\newblock URL \url{https://doi.org/10.1038/s43588-025-00902-w}.

\end{thebibliography}


\newpage

\appendix

\section{HELiX Technical Details}

\subsection{Model Architecture}

The policy model underlying HELiX is a graph transformer $f(s) \in \mathbb{R}^D$ operating on states $s = (m, r, F)$, where $m$ is a molecule, $r$ is a reaction template, and $F$ is a set of fragments. The model outputs a $D$-dimensional state embedding. The molecule $m$ is represented as a molecular graph. Reaction templates $r$ are encoded as one-hot vectors over the template library. Each fragment in $F$ is represented by an ECFP fingerprint, linearly embedded and summed to obtain a fixed-size fragment-set representation.

Let $A_F(s)$ denote the set of forward actions available from a state $s$, and $A_B(s)$ the corresponding set of backward actions. Both depend on the phase of $s$, which is determined by the decomposition and composition step counters. In the decomposition phase, the forward actions comprise reaction decomposition, fragment removal, and template removal, and the backward actions are their inverses: reaction application, fragment addition, and template addition. In the composition phase the two sets are exchanged.

Both forward and backward policies are defined by a unified action-scoring formulation. Since each action from $s$ yields a distinct successor state, we index the policies by
action here. Actions are scored and normalized using a softmax operator $\sigma^k$ over a length-$k$ logit vector $\mathbf{c} \in \mathbb{R}^k$:

\begin{equation}
    P_X(a_i \mid s) = \sigma^{|A_X(s)|}(\mathbf{c})_i, \qquad
    c_i = \phi\!\left(W_{A_X^*(s)} f(s)\right)^\top e_{A_X^*(s)}(a_i),
\end{equation}
 where $X \in \{F, B\}$ denotes the forward or backward policy, $A_X^*(s)$ denotes the type of action space $A_X(s)$, $\phi$ is the GeLU activation function, and $W_{A_X^*(s)}$ is a linear projection specific to the type of action space $A_X(s)$. Each action $a_i \in A_X(s)$ is embedded using an action-type-dependent embedding function $e_{A_X^*(s)}(\cdot)$. For reaction decomposition actions, where $a_i$ corresponds to a candidate predecessor tuple $(m', r, F)$, the action embedding is obtained by applying the backbone transformer, i.e., $e(a_i) = f(a_i)$. For reaction template addition, actions are embedded as one-hot vectors over the reaction template set. For fragment addition, the action embedding is formed by concatenating a one-hot fragment identifier with a linearly transformed ECFP fingerprint.

\subsection{Training Details}
\label{sec:training_details}

The forward policy $P_F$ is trained using the Trajectory Balance (TB) objective \citep{malkin2022}. The backward policy $P_B$ is trained jointly using the REINFORCE objective introduced in Section~\ref{sec:forward_and_backward_policies}. To mitigate the sparsity of the backward reward signal, half of the trajectories used to compute the REINFORCE loss are sampled via $P_F$ and stored in a replay buffer; such trajectories reach the initial hit by construction, providing a consistent source of positive reward signal. The forward policy additionally incorporates an exploitation penalty following \citet{gainski2025scalable} which discourages repeatedly visiting terminal states. The inner training loop is summarized in Algorithm~\ref{alg:helix_training}.

\begin{algorithm}[h]
\caption{HELiX Inner Training Loop}
\label{alg:helix_training}
\begin{algorithmic}[1]
\REQUIRE Hit molecule $H$, forward policy $P_F$, exploitation penalty $P_E$, backward policy $P_B$, reward $R:\mathcal{X}\to\mathbb{R}$
\STATE Initialize sampling policy $P_F' \leftarrow P_F \times P_E$
\FOR{each training iteration}
    \STATE Sample forward trajectories $\mathcal{T}_F$ from $P_F'$ starting from $x$
    \STATE Evaluate reward function $R$ on terminal states in $\mathcal{T}_F$
    \STATE Sample replay trajectories $\mathcal{T}_B$ from prioritized replay buffer using $P_B$
    \STATE Sample REINFORCE trajectories $\mathcal{T}_R$ from random replay buffer using $P_B$
    \STATE Update $P_F$ to minimize $\mathcal{L}_{\mathrm{TB}}$ over $\mathcal{T}_F \cup \mathcal{T}_B$
    \STATE Update $P_B$ to minimize $\mathcal{L}_{\mathrm{REINFORCE}}$ over $\mathcal{T}_F \cup \mathcal{T}_R$
    \STATE Update exploitation penalty counter $n(s, a)$
\ENDFOR
\RETURN
\end{algorithmic}
\end{algorithm}

\subsection{Bayesian Optimization}
\label{sec:appendix_bo}

We incorporate BO into HELiX training following the general procedure in Algorithm~\ref{alg:bo_loop}. We warm-start each run with 50 molecules sampled from a uniform forward policy, and run for $N = 20$ iterations with pool size $M = 1000$, using Top-$K$ subsampling with $q = 450$ in the first iteration and $q = 500$ thereafter, yielding a total budget of 10k oracle calls.

\begin{algorithm}[t]
\caption{Bayesian Optimization with GFlowNets}
\label{alg:bo_loop}
\begin{algorithmic}[1]
\STATE \textbf{Input:} Oracle $f: \mathcal{X} \rightarrow \mathbb{R}$, 
       initial dataset $\mathcal{D}_0$, BO iterations $N$, 
       pool size $M$, batch size $q \leq M$, acquisition function $\alpha$, 
       surrogate model $\hat{f}$
\STATE $\mathcal{D} \leftarrow \mathcal{D}_0$
\FOR{$i = 1$ to $N$}
    \STATE Fit surrogate $\hat{f}$ on $\mathcal{D}$
    \STATE Train GFlowNet policy $\pi_\theta$ with reward $R(x) = \alpha\big( x; \hat{f} \; \big)$
    \STATE Sample batch $\mathcal{B} \leftarrow \{x_j \sim \pi_\theta\}_{j=1}^{M}$, optionally select $q$ candidates via subsampling
    \STATE $\mathcal{D} \leftarrow \mathcal{D} \cup \{(x_j, f(x_j)\}_{j=1}^q$
\ENDFOR
\STATE \textbf{return} $\mathcal{D}$
\end{algorithmic}
\end{algorithm}

\paragraph{Surrogate model.} We use a Gaussian process (GP) surrogate model with the Tanimoto (MinMax) kernel over 4096-dimensional ECFP4 molecular fingerprints:
\begin{equation}
    k_{\mathrm{MinMax}}(x, x') = \frac{\sum_i \min(x_i, x'_i)}{\sum_i \max(x_i, x'_i)},
\end{equation}
fitted by maximizing the marginal log-likelihood. For efficiency, when $|\mathcal{D}| > 5000$, we subsample by retaining the top 2500 molecules by score and 2500 randomly selected molecules.

\paragraph{Acquisition Functions.} Given a surrogate $\hat{f}$ with posterior mean $\mu(x)$ and standard deviation $\sigma(x)$, we consider three standard acquisition functions: \textit{Expected Improvement} (EI) is defined as
\begin{equation}
    \alpha_{\mathrm{EI}}(x) = \mathbb{E}\big[\max(f(x) - f^*, 0)\big],
\end{equation}
where $f^* = \max_{x \in \mathcal{D}} f(x)$ is the best observed value. \textit{Upper Confidence Bound} (UCB) is defined as
\begin{equation}
    \alpha_{\mathrm{UCB}}(x) = \mu(x) + \gamma\, \sigma(x),
\end{equation}
where $\gamma \geq 0$ controls the exploration-exploitation tradeoff. As a special case, setting $\gamma = 0$ recovers the \textit{posterior mean} (PM) acquisition function. For the main HELiX experiments, we use UCB with $\gamma = 1$. For the analysis in Section~\ref{sec:experiments_bo}, we use $\gamma = 0.1$ and additionally study EI and PM, together with the effect of reward shaping via the power transformation $\tilde{y} = y^\beta$ described in Section~\ref{sec:methods_bo}.

\paragraph{Greedy Subsampling Methods.}
Given a batch of candidates $\mathcal{B} = \{x_j\}_{j=1}^{M}$, we consider three strategies for selecting the final set of $q < M$ candidates for oracle evaluation. \textit{Top-$K$} selects the $q$ candidates with the highest acquisition value. \textit{Thompson Sampling} (TS) draws an independent sample from the marginal posterior $\mathcal{N}\big(\mu(x_j), \sigma^2(x_j)\big)$ at each candidate and selects the maximizer, repeating this $q$ times without replacement. \textit{Multi-point UCB} (qUCB) selects $q$ candidates sequentially, maximizing a Monte Carlo estimate of the batch acquisition value over the joint posterior of the already-selected points together with the candidate~\citep{balandat_botorch_2020}.

\section{Experimental Details}
\label{sec:experimental_details}

\subsection{Initial hit generation}

\begin{figure}
    \centering
    \includegraphics[width=\linewidth]{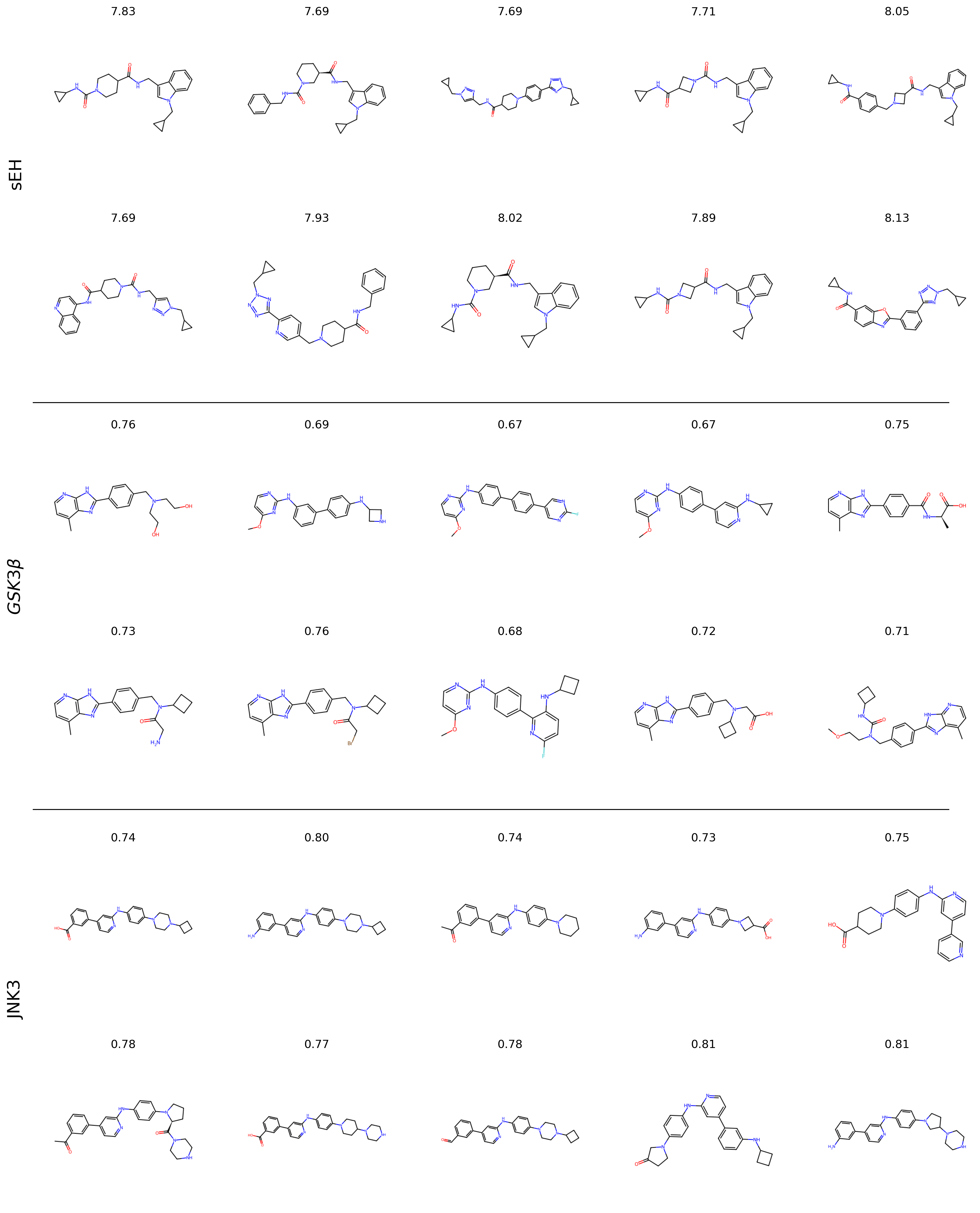}
    \caption{2D representations of initial hits generated by SCENT for sEH, GSK3$\beta$, and JNK3 with corresponding proxy scores.}
    \label{fig:initial_hits}
\end{figure}

For each oracle (sEH, GSK3$\beta$, JNK3), initial hit molecules are generated using a pretrained SCENT model that explores 320{,}000 molecules. From the resulting set, we retain the top 50\% of unique molecules by proxy score that satisfy the following drug-likeness criteria: QED $\geq 0.5$, SA score $\leq 3.0$, and molecular weight in the range $[300, 500]$ Da. To allow for at least 2 decomposition steps, we further restrict to molecules constructed with at least 3 reactions. The filtered set is then clustered greedily in order of proxy score, where cluster representatives are required to have pairwise ECFP6 Tanimoto similarity below 0.7 to existing centroids. The top 10 centroids by proxy score are taken as initial hits. SMILES strings and properties of the resulting compounds are shown in Table~\ref{tab:initial_hits}, and their 2D representations are shown in Figure~\ref{fig:initial_hits}.

\begin{table}[t]
\caption{Properties of the selected reference molecules generated by SCENT for sEH, GSK3$\beta$, and JNK3, separated respectively into three sections.}
\label{tab:initial_hits}
\centering
\footnotesize
\begin{tabular}{lrrrr}
\toprule
\textbf{Smiles} & \textbf{Proxy} & \textbf{QED} & \textbf{SA} & \textbf{MW} \\
\midrule

O=C(NCc1cn(CC2CC2)c2ccccc12)C1CCN(C(=O)NC2CC2)CC1 & 7.83 & 0.79 & 2.33 & 394 \\
O=C(NCc1cn(CC2CC2)c2ccccc12)[C@@H]1CCCN(C(=O)NCc2ccccc2)C1 & 7.69 & 0.571 & 2.73 & 444 \\
O=C(NCc1cn(CC2CC2)nn1)C1CCN(c2ccc(-c3nnn(CC4CC4)n3)cc2)CC1 & 7.69 & 0.521 & 2.7 & 461 \\
O=C(NC1CC1)C1CN(C(=O)NCc2cn(CC3CC3)c3ccccc23)C1 & 7.71 & 0.825 & 2.41 & 366 \\
O=C(NC1CC1)c1ccc(CN2CC(C(=O)NCc3cn(CC4CC4)c4ccccc34)C2)cc1 & 8.05 & 0.516 & 2.37 & 456 \\
O=C(Nc1ccnc2ccccc12)C1CCN(C(=O)NCc2cn(CC3CC3)nn2)CC1 & 7.69 & 0.622 & 2.46 & 433 \\
O=C(NCc1ccccc1)C1CCN(Cc2ccc(-c3nnn(CC4CC4)n3)nc2)CC1 & 7.93 & 0.59 & 2.4 & 431 \\
O=C(NCc1cn(CC2CC2)c2ccccc12)[C@@H]1CCCN(C(=O)NC2CC2)C1 & 8.02 & 0.79 & 2.8 & 394 \\
O=C(NCc1cn(CC2CC2)c2ccccc12)C1CN(C(=O)NC2CC2)C1 & 7.89 & 0.825 & 2.39 & 366 \\
O=C(NC1CC1)c1ccc2nc(-c3cccc(-c4nnn(CC5CC5)n4)c3)oc2c1 & 8.13 & 0.532 & 2.49 & 400 \\
\midrule
Cc1ccnc2[nH]c(-c3ccc(CN(CCO)CCO)cc3)nc12 & 0.76 & 0.615 & 2.49 & 326 \\
COc1ccnc(Nc2cccc(-c3ccc(NC4CNC4)cc3)c2)n1 & 0.69 & 0.636 & 2.27 & 347 \\
COc1ccnc(Nc2ccc(-c3ccc(-c4cnc(F)nc4)cc3)cc2)n1 & 0.67 & 0.520 & 2.33 & 373 \\
COc1ccnc(Nc2ccc(-c3ccnc(NC4CC4)c3)cc2)n1 & 0.67 & 0.713 & 2.24 & 333 \\
Cc1ccnc2[nH]c(-c3ccc(C(=O)N[C@H](C)C(=O)O)cc3)nc12 & 0.75 & 0.681 & 2.71 & 324 \\
Cc1ccnc2[nH]c(-c3ccc(CN(C(=O)CN)C4CCC4)cc3)nc12 & 0.73 & 0.741 & 2.54 & 349 \\
Cc1ccnc2[nH]c(-c3ccc(CN(C(=O)CBr)C4CCC4)cc3)nc12 & 0.76 & 0.638 & 2.62 & 413 \\
COc1ccnc(Nc2ccc(-c3nc(F)ccc3NC3CCC3)cc2)n1 & 0.68 & 0.631 & 2.47 & 365 \\
Cc1ccnc2[nH]c(-c3ccc(CN(CC(=O)O)C4CCC4)cc3)nc12 & 0.72 & 0.712 & 2.49 & 350 \\
COCCN(Cc1ccc(-c2nc3c(C)ccnc3[nH]2)cc1)C(=O)NC1CCC1 & 0.71 & 0.642 & 2.53 & 393 \\
\midrule
O=C(O)c1cccc(-c2ccnc(Nc3ccc(N4CCN(C5CCC5)CC4)cc3)c2)c1 & 0.74 & 0.584 & 2.23 & 428 \\
Nc1cccc(-c2ccnc(Nc3ccc(N4CCN(C5CCC5)CC4)cc3)c2)c1 & 0.8 & 0.607 & 2.24 & 399 \\
CC(=O)c1cccc(-c2ccnc(Nc3ccc(N4CCCCC4)cc3)c2)c1 & 0.74 & 0.586 & 2.04 & 371 \\
Nc1cccc(-c2ccnc(Nc3ccc(N4CC(C(=O)O)C4)cc3)c2)c1 & 0.73 & 0.602 & 2.2 & 360 \\
O=C(O)C1CCN(c2ccc(Nc3cc(-c4cccnc4)ccn3)cc2)CC1 & 0.75 & 0.697 & 2.16 & 374 \\
CC(=O)c1cccc(-c2ccnc(Nc3ccc(N4CCC[C@H]4C(=O)N4CCNCC4)cc3)c2)c1 & 0.78 & 0.53 & 2.95 & 469 \\
O=C(O)c1cccc(-c2ccnc(Nc3ccc(N4CCC(N5CCNCC5)CC4)cc3)c2)c1 & 0.77 & 0.515 & 2.43 & 457 \\
O=Cc1cccc(-c2ccnc(Nc3ccc(N4CCN(C5CCC5)CC4)cc3)c2)c1 & 0.78 & 0.579 & 2.32 & 412 \\
O=C1CCN(c2ccc(Nc3cc(-c4cccc(NC5CCC5)c4)ccn3)cc2)C1 & 0.81 & 0.595 & 2.39 & 398 \\
Nc1cccc(-c2ccnc(Nc3ccc(N4CCC(N5CCNCC5)C4)cc3)c2)c1 & 0.81 & 0.555 & 2.89 & 414\\
\bottomrule
\end{tabular}
\end{table}

\subsection{Trajectory settings}
For all experiments, we fix the maximum number of decomposition steps to 2 and allow a reaction budget of 1. Since initial hits are generated using 3 reaction steps, this ensures that sampled molecules share at least one reaction step with the initial hit while keeping molecular weight within a reasonable range.

\subsection{Baseline Methods}
\label{sec:baselines}

\paragraph{REINVENT4.}
We use the Mol2Mol component of REINVENT4 with a prior trained to generate molecules with ECFP4 Tanimoto similarity of at least 0.5 to the input molecule. All other generation parameters were kept at their default settings.

\paragraph{SynFormer-ED.}
We pre-trained a SynFormer-ED model on approximately 220 million data points, fewer than the original model due to the limited size of our chemical library, using a batch size of 512. To enable conditional generation, we modified the RL fine-tuning script originally designed for SynFormer-D to re-incorporate the trained encoder, updating only the decoder parameters with a batch size of 64. All other settings were kept at their default values, with both the prior and replay buffer enabled.

\paragraph{TRACER.}
We use the pre-trained Transformer and GCN weights provided by the authors, trained on the USPTO dataset. Since constructing a comparable dataset based on our specific library was not feasible, we treat TRACER as a black box where reaction templates and building blocks cannot be modified. MCTS was conducted using all default configurations.

\paragraph{SynGA.}
SynGA is a genetic algorithm that operates directly over synthesis routes. We use the same building block library and reaction templates as HELiX, restricting synthesis routes to at most 4 reaction steps to ensure fair comparison. The initial population of 100 molecules is seeded with the cluster's initial hit, with the remainder sampled randomly from the library. We use a population size of 500 with 5 offspring per generation, inverse-rank parent sampling, crossover probability 0.8, and mutation probability 0.5. The sEH proxy score is used directly as the fitness function. All other settings are kept at their defaults.

\paragraph{ReaSyn.} We used the provided pre-trained model to sample 10k molecules for each hit with all default settings.

\paragraph{SCENT.}
We evaluate SCENT within the same BO loop as HELiX, using UCB with $\gamma = 1$ and identical hyperparameters. Since SCENT is designed for \textit{de novo} exploration rather than local hit expansion, it serves as a baseline to highlight the necessity of local exploration. Trajectory length was limited to 4 reaction steps following the reaction budget in HELiX.

\subsection{Evaluation coverage}
\label{sec:coverage}

Ten initial hits were selected per target (Table~\ref{tab:initial_hits}), and each method was evaluated on multiple initial hits. All reported values are means and standard deviations across independent trials. However, coverage is not uniform across all methods and oracles. The number of independent trials for each method and oracle is reported in Table~\ref{tab:n_trials}. In no case were results filtered or hand-chosen. Reduced coverage reflects runs that did not complete, or the computational cost of the method; ReaSyn in particular was not evaluated on GSK3$\beta$ or JNK3 for this reason.

\begin{table}[h]
\caption{Number of initial hits on which each method was evaluated. Dashes indicate methods not evaluated on the corresponding target.}
\label{tab:n_trials}
\centering
\small
\setlength{\tabcolsep}{4pt}
\begin{tabular}{lcccccccc}
\toprule
Target & Uniform & SCENT & SynFormer-ED & REINVENT4 & TRACER & SynGA & ReaSyn & HELiX \\
\midrule
sEH          & 3  & 10 & 10 & 10 & 10 & 10 & 3  & 10 \\
GSK3$\beta$  & -- & -- & 10 & 10 & 10 & 3  & -- & 8  \\
JNK3         & -- & -- & 10 & 10 & 10 & 3  & -- & 5  \\
\bottomrule
\end{tabular}
\end{table}

\section{Library Details}
\label{sec:library_details}

\paragraph{SMALL Library.}
Our primary library follows the SMALL setting of RGFN \citep{koziarski2024}, consisting of 418 building blocks and 112 reaction templates. This library is used for all main experiments reported in Section~\ref{sec:experiments}.

\paragraph{CONCAT Library.}
To evaluate HELiX across larger fragment libraries, we construct the CONCAT library by taking the union of the SMALL building blocks with the LARGE building block set 
from SCENT \citep{gainski2025scalable}, which is drawn from the Enamine REAL database, yielding 128,154 total fragments. The reaction templates are kept identical to the SMALL setting. This ensures that any hit molecule decomposable under SMALL is also decomposable under CONCAT, allowing us to use the same reference molecules across library settings and attribute performance differences solely to the expanded fragment set. For the scaling experiments in Appendix~\ref{sec:ablations_library}, we construct intermediate libraries (concat\_2k through concat\_64k) by taking the union of the SMALL building blocks with a randomly sampled subset of the remaining CONCAT fragments until the desired library size is reached.

\paragraph{Enumerated Library.}
For the controlled BO analysis in Section~\ref{sec:experiments_bo}, we construct a small enumerated library by recursively applying a subset of reaction templates to a small set of building blocks, yielding a fixed candidate pool of approximately 30k molecules. This setting allows direct comparison between pool-based sampling and GFlowNet-based sampling over an identical design space.

\begin{figure}
    \centering
    \includegraphics[width=\linewidth]{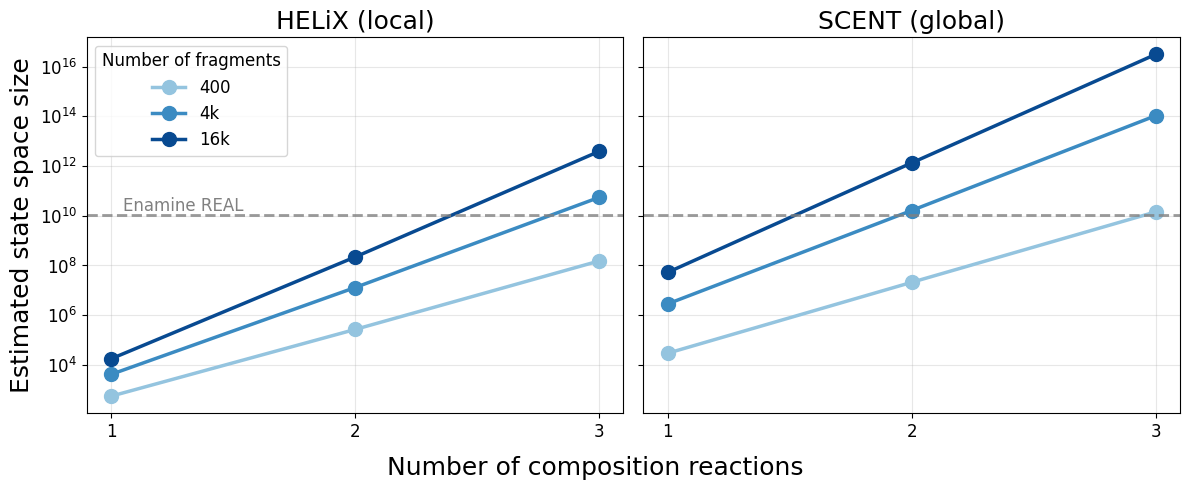}
    \caption{Estimated state space size for HELiX (local) and SCENT (global) as a function of the number of forward composition steps for fragment library sizes of 400, 4k, and 16k. The number of reaction templates is held constant. The Enamine REAL Space is shown as a reference. Despite being constrained to a local neighborhood, HELiX operates over a non-trivially large chemical space.}
    \label{fig:state_space}
\end{figure}

\paragraph{State Space Estimation.}
Following the methodology of \citet{koziarski2024}, we estimate the size of the chemical space accessible to HELiX as a function of fragment library size and trajectory length. We fix the number of decomposition steps to two and vary the number of forward composition steps from one to three. For reference, we perform the same estimation for SCENT. As shown in Figure~\ref{fig:state_space}, HELiX operates 2--4 orders of magnitude below SCENT at equivalent library size and depth, reflecting its constrained local search space. Despite this, the accessible space remains non-trivially large, with the Enamine REAL Space as a practical reference.

\section{Additional results}
\label{sec:additional_results}

\subsection{Results on sEH at Alternative Reward Threshold}
\label{app:seh_threshold}

Table~\ref{tab:seh_high} reports results on the sEH proxy at an alternative reward threshold ($> 8$). HELiX again achieves the highest number of modes and scaffolds.

\begin{table}[h]
\caption{Comparison to baselines on the sEH proxy at an alternative reward threshold. All metrics are computed over analogs of the initial hit (ECFP4 Tanimoto similarity $\geq 0.4$). Values are reported as mean $\pm$ standard deviation across initial hits; the number of hits evaluated per method is given in Appendix~\ref{sec:coverage}.}
\label{tab:seh_high}
\centering
\begin{tabular}{lcc}
\toprule
Method & Modes $> 8$ & Scaffolds $> 8$ \\
\midrule
Uniform   & $3 \pm 1$          & $8 \pm 1$            \\
SCENT     & $39 \pm 40$  & $150 \pm 172$  \\
SynFormer-ED & $49 \pm 42$ & $275 \pm 188$ \\
REINVENT4 & $5 \pm 4$    & $8 \pm 8$      \\
TRACER    & $3 \pm 3$    & $4 \pm 5$      \\
SynGA     & $7 \pm 2$          & $23 \pm 11$            \\
ReaSyn    & $1 \pm 1$          & $2 \pm 3$            \\
HELiX     & $\mathbf{68 \pm 29}$ & $\mathbf{287 \pm 129}$ \\
\bottomrule
\end{tabular}
\end{table}

\subsection{Results on GSK3$\beta$}
\label{sec:gsk3b}

Table~\ref{tab:gsk3b} reports results on the GSK3$\beta$ proxy. HELiX again achieves the highest number of local modes and scaffolds, as well as the highest success rate, while REINVENT4 achieves the highest top-10 and top-1 scores on this task.

\begin{table*}[t]
\caption{Comparison to baselines on GSK3$\beta$ proxy. Values are reported as mean $\pm$ standard deviation across initial hits; the number of hits evaluated per method is given in Appendix~\ref{sec:coverage}.}
\label{tab:gsk3b}
\centering
\vspace{6pt}
\small
\setlength{\tabcolsep}{3pt}
\begin{tabular}{lcccccccc}
\toprule
\multirow{2}{*}{Method} & \multirow{2}{*}{Analog} & \multirow{2}{*}{Success} & \multicolumn{2}{c}{Modes} & \multicolumn{2}{c}{Scaffolds} & \multirow{2}{*}{Top-10} & \multirow{2}{*}{Top-1} \\[3pt]
& & & $(> .75)$ & ($> .8)$ & $(> .75)$ & $(> .8)$ & & \\
\midrule
SynFormer-ED & .43 \std{.30} & .06 \std{.05} & 52 \std{60} & 7 \std{9} & 34 \std{29} & 2 \std{3} & .764 \std{.07} & .787 \std{.065} \\
REINVENT4 & \textbf{.81} \std{.04} & .13 \std{.09} & 255 \std{111} & 78 \std{26} & 150 \std{56} & 45 \std{28} & \textbf{.889} \std{.021} & \textbf{.923} \std{.02} \\
TRACER & .79 \std{.07} & .09 \std{.06} & 93 \std{44} & 20 \std{13} & 80 \std{58} & 19 \std{17} & .843 \std{.028} & .856 \std{.029} \\
SynGA & .13 \std{.07} & .05 \std{.03} & 73 \std{75} & 34 \std{43} & 93 \std{62} & 32 \std{39} & .834 \std{.056} & .853 \std{.054} \\
HELiX & .61 \std{.21} & \textbf{.35} \std{.16} & \textbf{747} \std{409} & \textbf{298} \std{223} & \textbf{556} \std{135} & \textbf{238} \std{131} & .88 \std{.031} & .891 \std{.031} \\
\bottomrule
\end{tabular}
\end{table*}

\subsection{Results on JNK3}
\label{sec:jnk3}

Table~\ref{tab:jnk3} reports results on the JNK3 proxy. HELiX achieves the highest analog rate and success rate, as well as the highest number of local modes and scaffolds, while REINVENT4 achieves the highest top-10 and top-1 scores on this task.

\begin{table*}[t]
\caption{Comparison to baselines on JNK3 proxy. Values are reported as mean $\pm$ standard deviation across initial hits; the number of hits evaluated per method is given in Appendix~\ref{sec:coverage}.}
\label{tab:jnk3}
\centering
\vspace{6pt}
\small
\setlength{\tabcolsep}{3pt}
\begin{tabular}{lcccccccc}
\toprule
\multirow{2}{*}{Method} & \multirow{2}{*}{Analog} & \multirow{2}{*}{Success} & \multicolumn{2}{c}{Modes} & \multicolumn{2}{c}{Scaffolds} & \multirow{2}{*}{Top-10} & \multirow{2}{*}{Top-1} \\[3pt]
& & & $(> .75)$ & ($> .8)$ & $(> .75)$ & $(> .8)$ & & \\
\midrule
SynFormer-ED & .64 \std{.10} & .09 \std{.04} & 194 \std{37} & 74 \std{16} & 616 \std{215} & 236 \std{94} & .884 \std{.012} & .903 \std{.021} \\
REINVENT4 & .89 \std{.03} & .17 \std{.09} & 515 \std{128} & 235 \std{93} & 879 \std{281} & 402 \std{195} & \textbf{.928} \std{.016} & \textbf{.949} \std{.022} \\
TRACER & .87 \std{.06} & .05 \std{.06} & 71 \std{28} & 21 \std{8} & 142 \std{64} & 35 \std{18} & .856 \std{.02} & .873 \std{.023} \\
SynGA & .30 \std{.10} & .08 \std{.02} & 162 \std{2} & 87 \std{7} & 602 \std{128} & 371 \std{83} & .911 \std{.015} & .92 \std{.016} \\
HELiX & \textbf{.92} \std{.07} & \textbf{.53} \std{.11} & \textbf{1005} \std{302} & \textbf{645} \std{278} & \textbf{4505} \std{655} & \textbf{3063} \std{739} & .911 \std{.017} & .92 \std{.019} \\
\bottomrule
\end{tabular}
\end{table*}

\subsection{Properties of Generated Molecules}
\label{sec:synth_metrics}

\begin{figure}
    \centering
    \includegraphics[width=\linewidth]{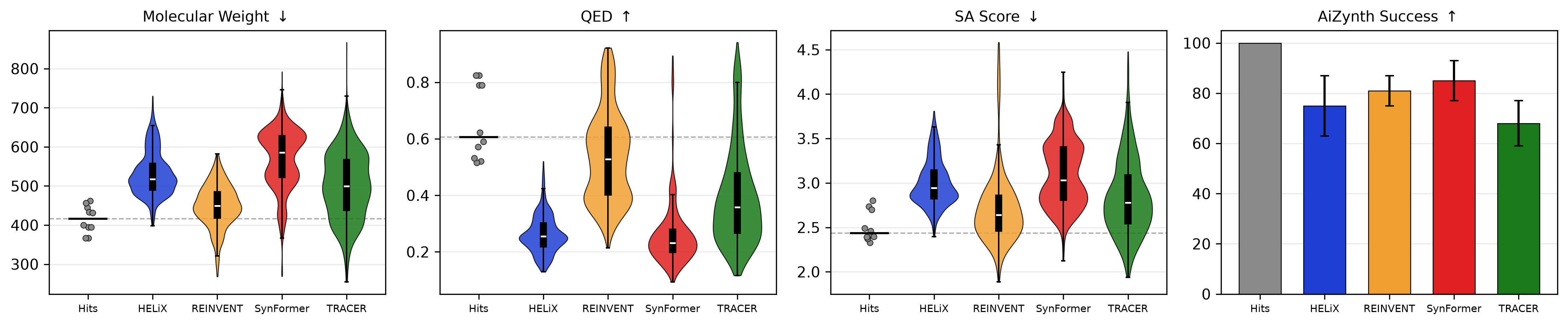}
    \caption{Distributions of molecular weight, QED, SA score, and AiZynthFinder retrosynthesis success rate across the top 100 molecules generated per initial hit (1000 molecules total per method). Arrows indicate the direction of improvement for each metric. The ten initial hits are shown in grey for reference, with the dashed line indicating their median; AiZynth success rate for the hits is 100\%.}
    \label{fig:mol_properties}
\end{figure}

Figure~\ref{fig:mol_properties} shows the distributions of molecular weight, QED, SA score, and AiZynthFinder (AZF) retrosynthesis success rate across the top 100 molecules generated per initial hit (1000 molecules total per model). The properties of the ten initial hits are shown in grey for reference. All four methods generate molecules that are heavier, less drug-like, and harder to synthesize than the initial hits. Median molecular weight rises from 416 Da to between 450 and 586 Da, median QED falls from .61 to between .23 and .53, and median SA score rises from 2.44 to between 2.64 and 3.03. REINVENT4 stays closest to the hits, consistent with its high analog rate and its tendency to make only small modifications to the starting molecule. Among the two methods that share a building-block library, HELiX remains closer to the hits than SynFormer-ED on all three properties. We emphasize that none of these properties enters the reward: HELiX optimizes the oracle alone. Incorporating drug-likeness directly into a multi-objective reward would address this, which we note as a limitation in Section~\ref{sec:conclusion}.

\paragraph{Drug-likeness metrics.}
REINVENT4 and TRACER exhibit better QED and SA scores than HELiX and SynFormer-ED. However, we note that these metrics are heavily biased toward the training data of these models: QED is derived from the ChEMBL DrugBank store of approved drugs, and SA score penalizes fragments that are rare in PubChem. Since REINVENT4 Mol2Mol is trained directly on ChEMBL and PubChem, and TRACER is trained on USPTO reactions which substantially overlaps with both, these models are biased toward generating compounds well-represented in existing drug-like datasets. HELiX and SynFormer-ED use the same reaction templates and building blocks, are not pretrained on drug-like corpora, and are directly comparable to each other. Among these two, HELiX achieves slightly better MW, QED, and SA score on average.

\paragraph{AiZynthFinder retrosynthesis success rate.}
Interestingly, SynFormer-ED achieves the highest AZF success rate despite having highly similar MW, QED, and SA distributions to HELiX. We find that this discrepancy is explained by proximity to the initial hit: the molecules generated by HELiX have mean Tanimoto similarity $0.42 \pm 0.03$ to the hit, while those generated by SynFormer-ED are closer to the hit at $0.49 \pm 0.12$. AZF was further able to retrosynthesize the 10 initial hits which were generated to have highly drug-like properties (see Section~\ref{sec:experimental_details}).

Since each of the 10 initial hits was successfully retrosynthesized by AZF, models that stay closer to the hit tend to inherit this property. To control for this, we perform a per-cluster proximity analysis: for each initial hit, we take the higher of the two models' mean similarities as a threshold and filter each model's molecules to those above it. Table~\ref{tab:azf_proximity} summarizes the per-cluster thresholds and resulting sample sizes. After this control, we find that HELiX and SynFormer-ED achieve nearly identical AZF success rates of 86.5\% and 86.6\%, respectively, confirming that the gap is driven by proximity rather than 
synthesizability.

\begin{table}[h]
\caption{Per-cluster proximity-controlled AZF analysis. Bold values indicate the 
threshold used for filtering (higher of the two cluster means). AZF success rates 
are computed on the filtered subsets.}
\label{tab:azf_proximity}
\centering
\begin{tabular}{ccccc}
\toprule
Hit & HELiX mean sim & SynFormer mean sim & HELiX $n$ & SynFormer $n$ \\
\midrule
1  & 0.388 & \textbf{0.594} & 0   & 55  \\
2  & 0.382 & \textbf{0.615} & 1   & 47  \\
3  & \textbf{0.411} & 0.366 & 40  & 34  \\
4  & 0.409 & \textbf{0.545} & 7   & 56  \\
5  & 0.477 & \textbf{0.534} & 19  & 43  \\
6  & \textbf{0.400} & 0.299 & 38  & 7   \\
7  & \textbf{0.456} & 0.315 & 34  & 5   \\
8  & 0.402 & \textbf{0.628} & 3   & 53  \\
9  & 0.394 & \textbf{0.578} & 0   & 52  \\
10 & \textbf{0.449} & 0.393 & 43  & 35  \\
Total & --- & --- & 185 & 387 \\
\midrule
AZF success rate & \multicolumn{2}{c}{86.5\% (HELiX)} 
                 & \multicolumn{2}{c}{86.6\% (SynFormer)} \\
\bottomrule
\end{tabular}
\end{table}

\subsection{Non-synthesizable Reference Molecules}
\label{sec:non_synth_reference}

It is possible that the initial hit compound is not compatible with the building blocks and reaction templates of HELiX, making it impossible to decompose and perform local exploration. To address this, we combine our method with ChemProjector, a recent method which proposes to ``project'' non-synthesizable molecules onto the nearest synthetic tree composed of fragments and reactions available in a given library \citep{luo2024projecting}. We generated 50 analogs of a known sEH inhibitor with ChemProjector. We then clustered the resulting molecules and took the top 5 according to their sEH proxy score. Table~\ref{tab:projection} shows the results of local exploration around these analogs. We note that, while the metrics are comparatively low, the output molecules from ChemProjector had relatively lower scores than the initial hits generated by SCENT. Thus, we also report the improvement upon the initial hit.

\begin{table*}[t]
\caption{Evaluation of HELiX combined with ChemProjector for non-synthesizable reference molecules.}
\label{tab:projection}
\vspace{4pt}
\centering
\begin{tabular}{lcccc}
\toprule
\textbf{Method} & \textbf{Modes} & \textbf{Top-10 Mean} & \textbf{Top-1} & \textbf{Improvement} \\
\midrule
HELiX              & 109 & 7.99 & 8.09 & 0.23 \\
HELiX + ChemProjector         & 17 & 7.71 & 7.75 & 1.64  \\
\bottomrule
\end{tabular}
\end{table*}

\subsection{SETDB1 Optimization Results}
\label{sec:setdb1_full}

Table~\ref{tab:setdb1_bo} reports the full results for the constrained-budget experiment in Section~\ref{sec:experiments_virtual_screening}, including Top-1 scores.

\begin{table}[h]
\caption{Optimization of SETDB1 docking scores under a budget of 1k oracle calls. Values are
AutoDock Vina scores (kcal/mol) over generated analogs; lower is better. Docking scores of
the initial hits are shown in Figure~\ref{fig:qualitative_binders}.}
\label{tab:setdb1_bo}
\centering
\begin{tabular}{lcccc}
\toprule
\multirow{2}{*}{Method} & \multicolumn{2}{c}{Hit 0} & \multicolumn{2}{c}{Hit 1} \\[3pt]
& Top-10 & Top-1 & Top-10 & Top-1 \\
\midrule
REINVENT4      & -9.47  & -10.10 & -10.68 & -11.30 \\
HELiX w/o BO   & -10.01 & -10.70 & -10.13 & -10.60 \\
HELiX          & \textbf{-10.34} & \textbf{-11.20} & \textbf{-12.93} & \textbf{-13.60} \\
\bottomrule
\end{tabular}
\vspace{-10pt}
\end{table}

\subsection{Physical validity of Docked Compounds}
\label{sec:posebusters}

We evaluate the physical validity of all 600 candidates resulting from virtual screening in Section~\ref{sec:experiments_virtual_screening} using PoseBusters \citep{buttenschoen2024posebusters}, which assesses geometric and chemical validity of docked poses. Of the 600 candidates, 82 could not be evaluated due to a known conversion issue from docking pose format to SDF with obabel. Of the remaining 518 molecules, 517 pass all PoseBusters checks (99.8\%), confirming that HELiX generates physically valid, well-docked candidates.

\subsection{Evaluation under a Held-Out Oracle}
\label{sec:heldout_oracle}
 
A concern when optimizing against a learned or simulated oracle is that high scores may reflect exploitation of artifacts specific to that scoring function rather than genuine improvement in binding. To test this, we re-evaluate generated candidates with an independent scoring function that is never used during optimization.
 
We use the SETDB1 setting of Section~\ref{sec:experiments_virtual_screening}, optimizing each of the two experimentally validated hits against AutoDock Vina under a reduced budget of 1k oracle calls. We then rescore the top-50 molecules discovered by each method using GNINA \citep{mcnutt2021gnina}, a learned scoring function trained independently of the Vina objective and therefore serving as a held-out oracle. We compare against REINVENT4, the strongest local baseline in this setting.
 
\begin{table}[h]
\caption{Held-out oracle evaluation on SETDB1. Mean GNINA CNN affinity over the top-50
molecules discovered by each method under a 1k-call AutoDock Vina budget. Higher is better.}
\label{tab:heldout_oracle}
\centering
\begin{tabular}{lcc}
\toprule
Method & Hit 0 & Hit 1 \\
\midrule
REINVENT4 & 5.91 & 5.51 \\
HELiX     & \textbf{6.29} & \textbf{6.23} \\
\bottomrule
\end{tabular}
\end{table}
 
HELiX attains higher mean CNN affinity than REINVENT4 on both hits, indicating that its improved docking scores transfer to an independent oracle rather than reflecting exploitation of Vina. We note that the rank correlation between the Vina objective and the held-out CNN affinity is low for both methods (Spearman $\rho < 0.3$), reflecting general disagreement between the two scoring functions; this disagreement is present in the molecules generated by both HELiX and REINVENT, and therefore does not favor either.

\section{Ablations}
\label{sec:ablations}

\subsection{Bayesian Optimization}
\label{sec:ablations_bo}

Table~\ref{tab:bo_ablation} shows the effect of incorporating BO into both HELiX and SCENT. BO consistently improves performance across both models for both analogs and unconstrained candidates, with the largest gains in local mode discovery. HELiX + BO achieves the highest local mode count and AUC Top-10 overall.

\begin{table}[h]
\caption{Effect of Bayesian optimization on HELiX and SCENT on the sEH proxy. Local and global modes are reported at 10k oracle calls; AUC Top-10 is computed globally.}
\label{tab:bo_ablation}
\centering
\begin{tabular}{lcccc}
\toprule
\multirow{2}{*}{Method} & \multicolumn{2}{c}{Modes} & \multirow{2}{*}{AUC Top-10} \\[3pt]
& Local & Global & \\
\midrule
SCENT           & 296  & 1539 & .906 \\
SCENT + BO      & 174  & 1962 & .942 \\
HELiX           & 542  & 1255 & .913 \\
HELiX + BO      & \textbf{933}  & 1486 & \textbf{.952} \\
\bottomrule
\end{tabular}
\end{table}

\paragraph{Predictive quality of the GP surrogate.}
Because the proxy oracles used in our main experiments are themselves learned functions, they may
present a setting that is favorable to a GP surrogate. To examine this, at each BO
iteration we fit the GP on the molecules observed so far and evaluate its rank correlation
(Spearman $\rho$) with the true oracle on the next acquired batch (Table~\ref{tab:gp_fit}).
 
\begin{table}[h]
\caption{Rank correlation (Spearman $\rho$) between the GP surrogate and the true oracle on the
next acquired batch, at selected BO iterations.}
\label{tab:gp_fit}
\centering
\begin{tabular}{lccc}
\toprule
Iteration & sEH & JNK3 & GSK3$\beta$ \\
\midrule
1    & .37 & .40 & .40 \\
2    & .62 & .76 & .60 \\
3    & .77 & .82 & .60 \\
5    & .85 & .86 & .70 \\
10   & .89 & .86 & .60 \\
15   & .86 & .90 & .84 \\
19   & .89 & .94 & .85 \\
\midrule
Mean & .81 & .84 & .72 \\
\bottomrule
\end{tabular}
\end{table}
 
The surrogate predicts these oracles well on average, and we agree that they are relatively easy to fit. We note, however, that predictive quality is poor in the early iterations, which is when sample efficiency matters most, and yet BO nonetheless yields gains there. Section~\ref{sec:experiments_virtual_screening} evaluates the same framework against a docking oracle.
 
\paragraph{BO ablations across all three oracles.}
Table~\ref{tab:bo_ablation_all} extends the ablation above to the GSK3$\beta$ and JNK3 oracles. The ``HELiX'' rows correspond to the HELiX rows of Tables~\ref{tab:baselines_seh}, \ref{tab:gsk3b} and \ref{tab:jnk3}, as the experimental conditions are identical, so these results are directly comparable to the baselines reported there.
 
\begin{table}[h]
\caption{Effect of Bayesian optimization on HELiX across all three oracles. Modes and scaffolds are reported at the lower reward threshold used for each oracle. AUC Top-10 is computed globally.}
\label{tab:bo_ablation_all}
\centering
\small
\setlength{\tabcolsep}{4pt}
\begin{tabular}{llccccc}
\toprule
Oracle & Method & Modes & Scaffolds & Top-10 & Top-1 & AUC Top-10 \\
\midrule
\multirow{2}{*}{sEH} & HELiX w/o BO & 542 \std{133} & 1872 \std{538} & 8.12 \std{.10} & 8.22 \std{.1} & .913 \\
& HELiX & \textbf{933} \std{227} & \textbf{4559} \std{1221} & \textbf{8.22} \std{.05} & \textbf{8.31} \std{.07} & \textbf{.952} \\
\midrule
\multirow{2}{*}{GSK3$\beta$} & HELiX w/o BO & 18 \std{30} & 9 \std{16} & .735 \std{.048} & .77 \std{.04} & .546 \\
& HELiX & \textbf{747} \std{409} & \textbf{556} \std{135} & \textbf{.88} \std{.031} & \textbf{.891} \std{.031} & \textbf{.827} \\
\midrule
\multirow{2}{*}{JNK3} & HELiX w/o BO & 18 \std{7} & 24 \std{6} & .816 \std{.003} & .86 \std{.01} & .606 \\
& HELiX & \textbf{1005} \std{302} & \textbf{4505} \std{655} & \textbf{.911} \std{.017} & \textbf{.92} \std{.019} & \textbf{.876} \\
\bottomrule
\end{tabular}
\end{table}
 
BO improves every metric on all three oracles. The effect is far more pronounced on GSK3$\beta$ and JNK3 than on sEH, where the non-BO variant already performs strongly. We take this as evidence that the performance of HELiX is attributable both to the local exploration framework and to the BO-guided training loop, with their relative contributions depending on the oracle.

\subsection{Pool Size in Greedy Subsampling}
\label{sec:ablations_pool_size}

\begin{figure}[h]
    \centering
    \includegraphics[width=.6\linewidth]{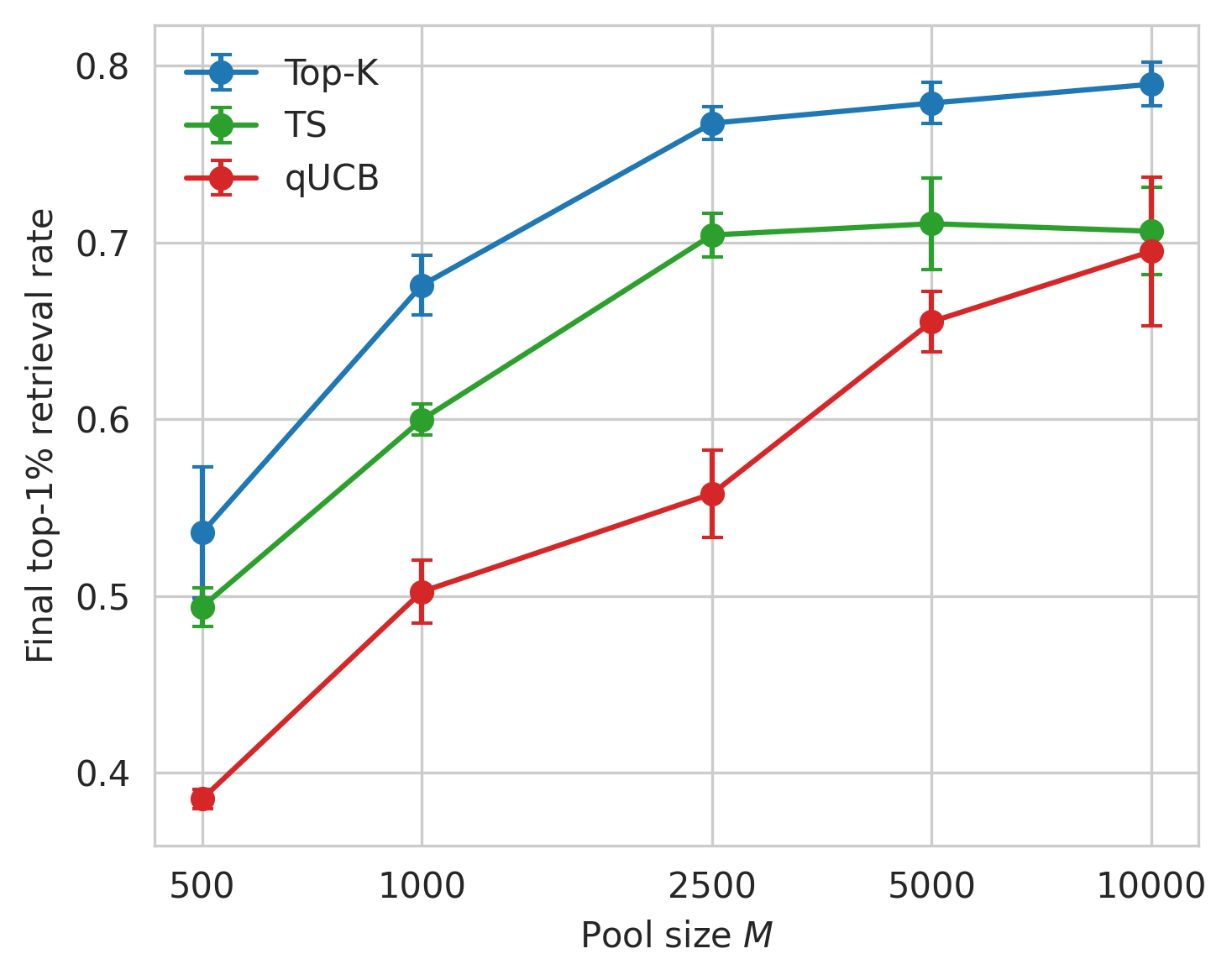}
    \caption{Final top-1\% retrieval rate after 1k oracle calls versus pool size $M$ for Top-$K$, TS, and qUCB subsampling with the pool-based sampler. Results are averaged over three seeds; error bars show $\pm 1$ standard deviation.}
    \label{fig:m_ablation_all}
\end{figure}

Figure~\ref{fig:m_ablation_all} reports the final top-1\% retrieval rate after 1k oracle calls for each greedy subsampling method under the pool-based sampler, across pool sizes $M \in \{500, 1\mathrm{k}, 2.5\mathrm{k}, 5\mathrm{k}, 10\mathrm{k}\}$. All runs follow the setting of Section~\ref{sec:experiments_bo}: 20 BO iterations with batch size $q = 50$ on the enumerated JNK3 library, sampling proportionally to PM. Performance improves with $M$ for all three methods, and Top-$K$ is best at every pool size.

\subsection{Analog Definition and Similarity Metrics}
\label{sec:ablations_sim}

\begin{figure}[h]
\centering
\includegraphics[width=\linewidth]{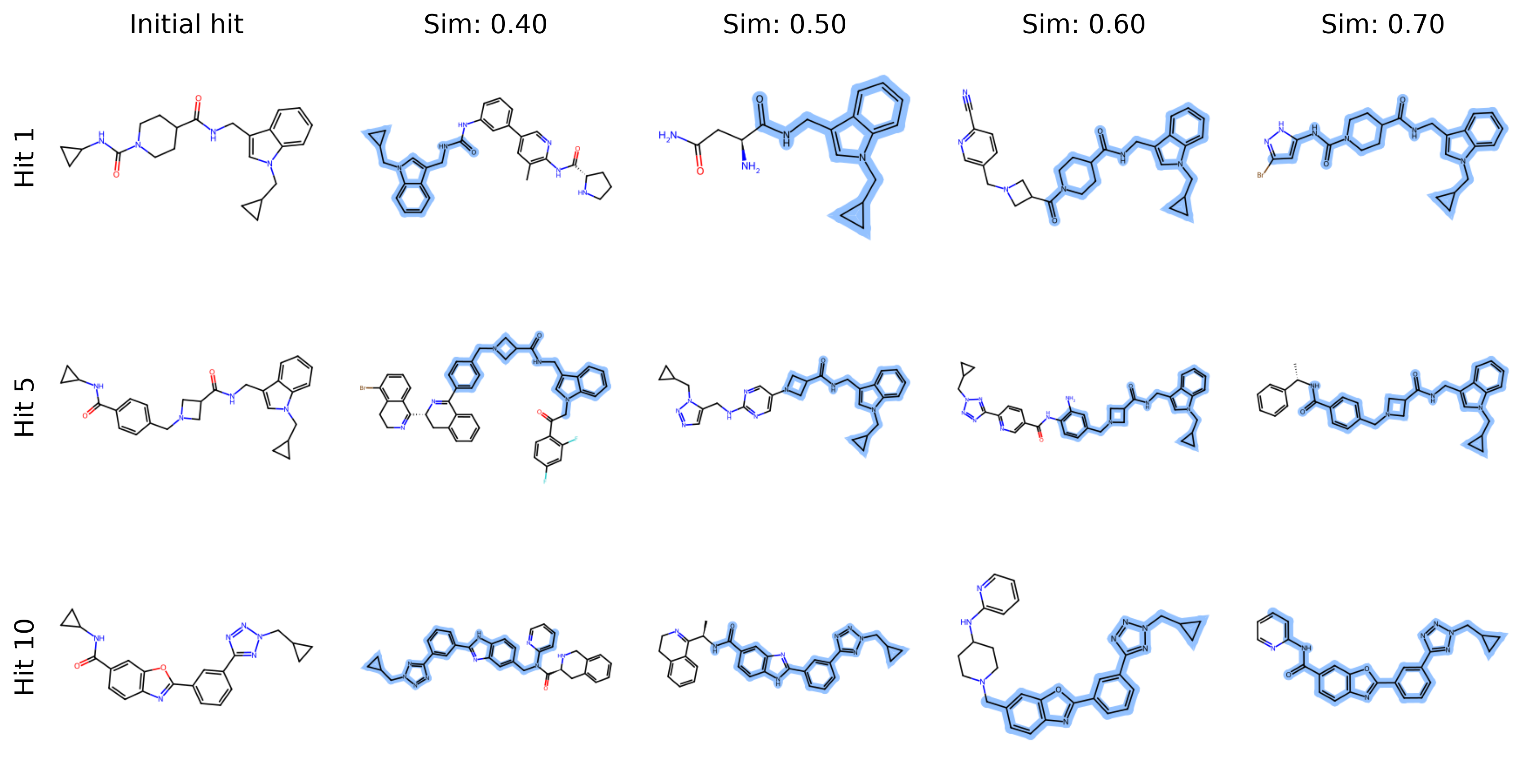}
\caption{Example analogs generated by HELiX at different ECFP4 Tanimoto similarity values to the initial hit. Each row corresponds to one initial hit, shown at left; each column shows a generated molecule at the indicated similarity, with the maximum common substructure shaded.}
\label{fig:analog_examples}
\end{figure}

\paragraph{Justification for the ECFP4 Tanimoto $\geq 0.4$ threshold.}
To characterize what this threshold means structurally, we compute the maximum common substructure (MCS) between each generated molecule and its initial hit, expressed as a fraction of the hit's heavy atoms, analyzed under different similarity thresholds (Table~\ref{tab:mcs_threshold}). At ECFP4 Tanimoto $\geq 0.4$, analogs retain a median 65\% of the hit's heavy atoms and differ from it by a median of 11 heavy atoms. The median building block in our library contains 10 heavy atoms, so this threshold corresponds to approximately a single building-block substitution---i.e., the granularity at which the search space is defined. Figure~\ref{fig:analog_examples} shows examples at each similarity level for three initial hits, with the common substructure shaded.
 
\begin{table}[h]
\caption{Structural interpretation of ECFP4 Tanimoto thresholds for HELiX on the sEH proxy. Median MCS is reported as a fraction of the initial hit's heavy atoms. ``Building blocks'' is the implied number of building-block substitutions, given a median library fragment size of 10 heavy atoms.}
\label{tab:mcs_threshold}
\centering
\begin{tabular}{lcccc}
\toprule
Threshold & Analog rate & Median MCS & Heavy atoms changed & Building blocks \\
\midrule
$\geq 0.3$ & .95 & .63 & 11 & 1.1 \\
$\geq 0.4$ & .65 & .65 & 11 & 1.1 \\
$\geq 0.5$ & .23 & .73 & 9  & 0.9 \\
$\geq 0.6$ & .04 & .80 & 6  & 0.6 \\
\bottomrule
\end{tabular}
\end{table}

\paragraph{Sensitivity to fingerprint representation.}
We recompute analog rate and success rate for all methods under ECFP4, ECFP6, FCFP4, atom-pair, and MACCS fingerprints at thresholds 0.3, 0.4, 0.5, and 0.6, giving 20 alternative analog definitions. HELiX achieves the highest success rate among all methods in 16 of these settings and SynFormer-ED in 4. The four exceptions all occur at thresholds 0.5--0.6, where analog rates fall below 10\% and comparisons are thus noisy, as discussed in the threshold ablation below.
 
\paragraph{Alternative similarity criteria.}
We further evaluate two definitions of similarity that are not based on molecular fingerprints: BRICS fragment similarity \citep{degen2008brics}, and Tversky similarity to the hit with $\alpha = 1, \beta = 0$, which measures the fraction of the hit contained in the analog and is insensitive to molecular size  \citep{tversky1977features}. For each measure we set the analog threshold such that the same fraction of molecules is admitted as under the original ECFP4 definition, pooling generated molecules across both methods so that analog definition does not depend on either method. This yields thresholds of .462 for BRICS similarity and .621 for Tversky similarity. Results are reported in Table~\ref{tab:alt_similarity}. Under both criteria HELiX retains higher analog rate and success rate, indicating that its performance is not specific to the ECFP4 similarity threshold.
 
\begin{table}[h]
\caption{Analog and success rates under similarity criteria that are not based on molecular fingerprints; evaluated on the sEH proxy. Thresholds are calibrated on the pooled distribution so that each criterion admits the same fraction of molecules as ECFP4 Tanimoto $\geq 0.4$.}
\label{tab:alt_similarity}
\centering
\begin{tabular}{lcccc}
\toprule
& \multicolumn{2}{c}{Analog rate} & \multicolumn{2}{c}{Success rate} \\
\cmidrule(lr){2-3} \cmidrule(lr){4-5}
Criterion & HELiX & SynFormer-ED & HELiX & SynFormer-ED \\
\midrule
BRICS & \textbf{.83} & .56 & \textbf{.17} & .08 \\
Tversky & \textbf{.79} & .51 & \textbf{.16} & .06 \\
\bottomrule
\end{tabular}
\end{table}
 
\paragraph{Scaffold retention versus scaffold diversity.}
For the hits considered here, the Bemis-Murcko scaffold \citep{bemis1996properties} is the entire molecule: these compounds are ring systems joined by short linkers, with no side chains. Scaffold similarity is therefore nearly identical to whole-molecule similarity ($\rho = .99$). We accordingly interpret the scaffold counts reported throughout this paper as the number of distinct high-reward analogs discovered, rather than as scaffold-level diversity in the medicinal-chemistry sense. This affects all methods equally and does not change the reported rankings. As a direct measure of structural retention, the median MCS with the hit is .65 for HELiX versus .50 for SynFormer-ED, and mean ring-system retention is .77 versus .63.
 
\paragraph{Sensitivity to molecular weight.}
Tanimoto similarity is sensitive to molecular weight, so we additionally stratify analog and success rates by molecular weight relative to the initial hit (Table~\ref{tab:mw_strat}). HELiX produces analogs within a similar MW range to the hit, whereas SynFormer-ED tends to produce larger molecules. Within groups of equivalent MW, HELiX achieves a substantially higher success rate, with the exception of $\Delta\mathrm{MW} > 150$, where SynFormer-ED produces a higher proportion of analogs. However, we note that these molecules are substantially larger than the initial hit.
 
\begin{table}[h]
\caption{Analog and success rates stratified by molecular weight relative to the initial hit, sEH proxy.}
\label{tab:mw_strat}
\centering
\begin{tabular}{lcc}
\toprule
$\Delta$MW (Da) & Analog rate (HELiX / SynFormer-ED) & Success rate (HELiX / SynFormer-ED) \\
\midrule
$(-50, 0]$   & \textbf{.83} / .51 & \textbf{.22} / .04 \\
$(0, 50]$    & \textbf{.73} / .60 & \textbf{.24} / .04 \\
$(50, 100]$  & \textbf{.74} / .65 & \textbf{.19} / .05 \\
$(100, 150]$ & \textbf{.70} / .69 & \textbf{.13} / .09 \\
$(150, 200]$ & .64 / \textbf{.74} & .11 / \textbf{.12} \\
$(200, 300]$ & .57 / \textbf{.77} & .08 / \textbf{.17} \\
\bottomrule
\end{tabular}
\end{table}
 
\paragraph{Threshold ablation.}
Tables~\ref{tab:sim_00},~\ref{tab:sim_05} and~\ref{tab:sim_06} report results at Tanimoto similarity thresholds of 0, 0.5 and 0.6 respectively. Without any locality constraint, SCENT performs best as expected---its unconstrained search space is a natural advantage in the global setting---while HELiX outperforms all non-de-novo baselines by a wide margin. At threshold 0.5, similar rankings hold to those in Table~\ref{tab:baselines_seh}: HELiX achieves the highest mode count, with SynFormer-ED competitive on scaffolds. At threshold 0.6, all methods approach zero modes and scaffolds due to the low analog rate in this regime, making comparisons noisy. This can in principle be addressed by reducing the number of decomposition steps in HELiX to constrain generation to a tighter neighborhood.

\begin{table}[h]
\caption{Baseline comparison on the sEH proxy with no similarity constraint.}
\label{tab:sim_00}
\centering
\begin{tabular}{lcccc}
\toprule
Method & Modes $>$ 7.2 & Scaffolds $>$ 7.2 & Top-10 Mean \\
\midrule
SCENT        & \textbf{1962 \std{179}} & \textbf{8000 \std{180}} & \textbf{8.32 \std{.03}} \\
SynFormer-ED & 897 \std{357}  & 4705 \std{578} & 8.26 \std{.10} \\
REINVENT4    & 215 \std{83}   & 434 \std{198}  & 7.99 \std{.10} \\
TRACER       & 120 \std{72}   & 288 \std{185}  & 7.97 \std{.13} \\
SynGA       & 309 \std{76}   & 1191 \std{355}  & 8.11 \std{.05} \\
ReaSyn       & 76 \std{67}   & 445 \std{587}  & 7.86 \std{.27} \\
HELiX        & 1486 \std{226} & 7222 \std{613} & 8.28 \std{.06} \\
\bottomrule
\end{tabular}
\end{table}

\begin{table}[h]
\caption{Baseline comparison on the sEH proxy at Tanimoto similarity threshold $\geq 0.5$.}
\label{tab:sim_05}
\centering
\begin{tabular}{lccccc}
\toprule
Method & Modes $>$ 7.2 & Scaffolds $>$ 7.2 & Top-10 Mean & Analog Rate \\
\midrule
SCENT        & 26 \std{29}   & 69 \std{89}     & 7.28 \std{2.56} & .007 \std{.01} \\
SynFormer-ED & 235 \std{69}  & \textbf{1696 \std{724}}  & 8.18 \std{.08} & .31 \std{.13} \\
REINVENT4    & 144 \std{64}  & 315 \std{148}   & 7.98 \std{.10} & \textbf{.53 \std{.15}} \\
TRACER       & 57 \std{49}   & 170 \std{149}   & 7.80 \std{.32} & .27 \std{.16} \\
SynGA       & 97 \std{41}   & 423 \std{223}  & 8.06 \std{.1} & .14 \std{.06} \\
ReaSyn       & 35 \std{37}   & 137 \std{186}  & 7.77 \std{.29} & .08 \std{.07} \\
HELiX        & \textbf{312 \std{149}} & 1499 \std{821}  & \textbf{8.18 \std{.06}} & .23 \std{.12} \\
\bottomrule
\end{tabular}
\end{table}

\begin{table}[h]
\caption{Baseline comparison on the sEH proxy at Tanimoto similarity threshold $\geq 0.6$.}
\label{tab:sim_06}
\centering
\begin{tabular}{lccccc}
\toprule
Method & Modes $>$ 7.2 & Scaffolds $>$ 7.2 & Top-10 Mean & Analog Rate \\
\midrule
SCENT        & 3 \std{4}    & 7 \std{10}     & 4.78 \std{4.12} & .001 \std{.001} \\
SynFormer-ED & \textbf{61 \std{20}}  & \textbf{490 \std{233}}  & 8.01 \std{.08} & .09 \std{.04} \\
REINVENT4    & 53 \std{30}  & 135 \std{82}   & 7.92 \std{.13} & \textbf{.18 \std{.07}} \\
TRACER       & 19 \std{14}  & 68 \std{66}    & 7.69 \std{.38} & .08 \std{.08} \\
SynGA       &30 \std{15}  & 129 \std{78} & 7.99 \std{.12} & .04 \std{.02} \\
ReaSyn       & 11 \std{12}   & 40 \std{51}  & 7.65 \std{.4} & .02 \std{.01} \\
HELiX        & 53 \std{33}  & 224 \std{178}  & \textbf{8.04 \std{.11}} & .04 \std{.003} \\
\bottomrule
\end{tabular}
\end{table}

\paragraph{On the generality of our definition.}
We emphasize that ``analog'' is a loose notion in general, and that no single similarity criterion is appropriate across all campaigns. Our threshold was chosen as a reasonable and widely used default, and the analyses above indicate that the performance of HELiX is not sensitive to this choice. We further note that the mechanism by which HELiX imposes locality---shared synthesis routes---does not rely on any similarity measure at all, and can be controlled directly by the user through the number of decomposition and forward reaction steps, or through the choice of fragment and reaction-template libraries.

\subsection{Library Size}
\label{sec:ablations_library}

Table~\ref{tab:concat} compares library-dependent methods on the CONCAT library (128k fragments). HELiX maintains a substantially larger lead over SynFormer-ED compared to the SMALL library setting, and SCENT discovers no high-reward analogs, further demonstrating the necessity of local exploration frameworks as the fragment library grows.

\begin{table}[h]
\caption{Comparison of library-dependent methods on the CONCAT library (128k fragments), sEH proxy. Local metrics are computed at Tanimoto similarity 
$\geq 0.4$.}
\label{tab:concat}
\centering
\begin{tabular}{lcccccc}
\toprule
& \multicolumn{2}{c}{Analogs} & \multicolumn{2}{c}{Global} \\[3pt]
Method & Modes $>$ 7.2 & Scaffolds $>$ 7.2 & Modes $>$ 7.2 & Scaffolds $>$ 7.2 \\
\midrule

SCENT        & 0             & 0             & \textbf{2301} & \textbf{8680} \\
SynFormer-ED & 469           & 2944          & 761  & 4383 \\
HELiX        & \textbf{1402} & \textbf{5606} & 2408 & 8845 \\
\bottomrule
\end{tabular}
\end{table}


\end{document}